\documentclass[twoside]{article}

\usepackage[accepted]{aistats2023}
\usepackage{color}

\usepackage{graphicx}
\usepackage{subcaption}
\usepackage{hyperref}
\usepackage{url}
\usepackage{booktabs}
\usepackage{multirow}
\def\ourss{BGAE\ }
\def\ours{BGAE}
\def\vourss{BVGAE\ }
\def\vours{BVGAE}

\def\oursFull{\textbf{B}arlow \textbf{G}raph \textbf{A}uto-\textbf{E}ncoder}

\def\voursFull{\textbf{B}arlow \textbf{V}ariational \textbf{G}raph \textbf{A}uto-\textbf{E}ncoder}

\usepackage{amsmath,amsfonts,bm}

\def\tabref#1{table~\ref{#1}}
\def\Tabref#1{Table~\ref{#1}}

\def\figref#1{figure~\ref{#1}}
\def\Figref#1{Figure~\ref{#1}}

\def\secref#1{section~\ref{#1}}

\def\eqref#1{equation~\ref{#1}}
\def\Eqref#1{Equation~\ref{#1}}

\def\1{\bm{1}}

\def\rvw{{\mathbf{w}}}

\def\rvz{{\mathbf{z}}}

\def\rmA{{\mathbf{A}}}

\def\rmD{{\mathbf{D}}}

\def\rmI{{\mathbf{I}}}

\def\rmS{{\mathbf{S}}}
\def\rmT{{\mathbf{T}}}

\def\rmX{{\mathbf{X}}}

\def\rmZ{{\mathbf{Z}}}

\DeclareMathAlphabet{\mathsfit}{\encodingdefault}{\sfdefault}{m}{sl}
\SetMathAlphabet{\mathsfit}{bold}{\encodingdefault}{\sfdefault}{bx}{n}

\def\gB{{\mathcal{B}}}
\def\gC{{\mathcal{C}}}

\def\gE{{\mathcal{E}}}

\def\gG{{\mathcal{G}}}

\def\gI{{\mathcal{I}}}

\def\gL{{\mathcal{L}}}

\def\gN{{\mathcal{N}}}

\def\gV{{\mathcal{V}}}

\def\sE{{\mathbb{E}}} 

\def\sR{{\mathbb{R}}}

\newcommand{\sigmoid}{\sigma}

\newcommand{\Ahat}{\mathbf{\hat{A}}}

\begin{document}

\twocolumn[

  \aistatstitle{Barlow Graph Auto-Encoder for Unsupervised Network Embedding}

  \aistatsauthor{ Rayyan Ahmad Khan \And Martin Kleinsteuber }

  \aistatsaddress{ Unite Network SE, Germany \\  TU Munich, Germany \\ rayyan.khan@tum.de  \And  Unite Network SE, Germany \\  TU Munich, Germany \\ martin.kleinsteuber@unite.eu}
]
\runningauthor{Khan, R.}

\begin{abstract}
  Network embedding has emerged as a promising research field for network analysis.
  Recently, an approach, named Barlow Twins, has been proposed for self-supervised learning in computer vision by applying the \textit{redundancy-reduction principle} to the embedding vectors corresponding to two distorted versions of the image samples.
  Motivated by this, we propose \oursFull, a simple yet effective architecture for learning network embedding.
  It aims to maximize the similarity between the embedding vectors of immediate and larger neighborhoods of a node, while minimizing the redundancy between the components of these projections.
  In addition, we also present the variational counterpart named \voursFull.
  We demonstrate the effectiveness of \ourss and \vourss in learning multiple graph-related tasks, i.e., link prediction, clustering, and downstream node classification, by providing extensive comparisons with several well-known techniques on eight benchmark datasets.
\end{abstract}

\section{Introduction}
Graphs are flexible data structures used to model complex relations in a myriad of real-world phenomena such as biological and social networks, chemistry, knowledge graphs, and many others~\cite{applications-social3,applications-bio1,applications-chem1,applications-others1,applications-survey1}.
Recent years have seen a remarkable interest in the field of graph analysis in general and unsupervised network embedding learning in particular.
Network embedding learning aims to project the graph nodes into a continuous low-dimensional vector space such that the structural properties and semantics of the network are preserved~\cite{embeddings-survey1,embeddings-survey2}.
The quality of the embedding is determined by the downstream tasks such as transductive node classification and clustering.

A variety of approaches have been proposed over the years for learning network embedding.
On one hand, we have techniques that aim to learn the network embedding by employing the proximity information which is not limited to the first-hop neighbors.
Such approaches include spectral, random-walk based and matrix-factorization based methods~\cite{deepwalk,node2vec,spectral-conv,m-nmf}.
On the other hand, most neural network based approaches focus on the structural information by limiting themselves to the immediate neighborhood~\cite{gcn,gat,gin}.
Intuitively, larger neighborhood offers richer information that should consequently help in learning better network embedding.
However, the neural network based approaches often yield better results compared to the spectral techniques etc., despite them being theoretically more elegant.
Recently, graph Diffusion~\cite{diffusion} has been proposed to enable a variety of graph based algorithms to make use of a larger neighborhood in the graphs with high homophily.
This is achieved by precomputing a graph diffusion matrix from the adjacency matrix, and then using it in place of the original adjacency matrix.
For instance, coupling this technique with graph neural networks (GNNs) enables them to learn from a larger neighborhood, thereby improving the network embedding learning.
However, replacing the adjacency matrix with the diffusion matrix deprives the algorithms from an explicit local view provided by the immediate neighborhood, and forces them to learn only from the global view presented by the diffusion matrix.
Such an approach can affect the performance of the learning algorithm especially in the graphs where the immediate neighborhood holds high significance.
This advocates the need to revisit the way in which the information in the multi-hop neighborhood is employed to learn the network embedding.
There exist some contrastive approaches like~\cite{dgi,contrastive1,grace} that can capture the information in larger neighborhood in the form of the summary vectors, and then learn network embedding by aiming to maximize the local-global mutual information between local node representations and the global summary vectors.
However, this information is captured in an \textit{implicit} manner in the sense that there is no objective function to ensure preservation of the information in the larger neighborhood.

In this work we adopt a novel approach for learning network embedding by simultaneously employing the information in the immediate as well as larger neighborhood in an \textit{explicit} manner.
This is achieved by learning concurrently from the adjacency matrix and the graph diffusion matrix.
To efficiently merge the two sources of information, we take inspiration from Barlow Twins, an approach recently proposed for unsupervised learning of the image embeddings by constructing the cross-correlation matrix between the outputs of two identical networks fed with distorted versions of image samples, and making it as close to the identity matrix as possible~\cite{barlow}.
Motivated by this, we propose an auto-encoder-based architecture named as \oursFull(\ours), along with its variational counterpart named as \voursFull(\vours).
Both \ourss and \vourss make use of the immediate as well as the larger neighborhood information to learn network embedding in an unsupervised manner while minimizing the redundancy between the components of the low-dimensional projections.
Our contribution is three-fold:
\begin{itemize}
    \item We propose a simple yet effective auto-encoder-based architecture for unsupervised network embedding, which \textit{explicitly} learns from both the immediate and the larger neighborhoods provided in the form of the adjacency matrix and the graph diffusion matrix respectively.
    \item Motivated by Barlow Twins, \ourss and \vourss aim to achieve stability towards distortions and redundancy-minimization between the components of the embedding vectors.
    \item We show the efficacy of our approach by evaluating it on link-prediction, transductive node classification and clustering on eight benchmark datasets.
    Our approach consistently yields promising results for all the tasks whereas the included competitors often under-perform on one or more tasks.
\end{itemize}
As detailed in the subsequent sections and derivations, it is non-trivial to efficiently merge the information from neighborhoods at different levels, as it needs a careful choice of the loss function and related architecture components.
\section{Related Work}
\subsection{Network Embedding}
Earlier work related to network embedding, such as GraRep~\cite{grarep}, HOPE~\cite{hope}, and M-NMF~\cite{m-nmf}, etc., employed matrix factorization based techniques.
Concurrently, some probabilistic models were proposed to learn network embedding by using random-walk based objectives.
Examples of such approaches include DeepWalk~\cite{deepwalk}, Node2Vec~\cite{node2vec}, and LINE~\cite{line}, etc.
Such techniques over-emphasize the information in proximity, thereby sacrificing the structural information~\cite{struc2vec}.
In recent years several graph neural network (GNN) architectures have been proposed as an alternative to matrix-factorization and random-walk based methods for learning the graph-domain tasks.
Some well-known examples of such architectures include graph convolutional network or GCN~\cite{gcn}, graph attention network or GAT~\cite{gat}, Graph Isomorphism Networks or GIN~\cite{gin}, and GraphSAGE~\cite{graph-sage}, etc.
This has allowed exploration of network embedding using GNNs~\cite{embeddings-survey1,embeddings-survey2}.
Such approaches include auto-encoder based (e.g., VGAE~\cite{vgae} and GALA~\cite{gala}), adversarial (e.g, ARVGA~\cite{arvga} and DBGANp~\cite{dbgan}), and contrastive techniques (e.g., DGI~\cite{dgi}, MVGRL~\cite{mvgrl} and GRACE~\cite{grace}), etc.

\subsection{Barlow Twins} \label{sec:related:barlow}
This approach has been recently proposed \cite{barlow} as a self-supervised learning (SSL) mechanism making use of redundancy reduction - a principle first proposed in neuroscience \cite{barlow-old}.
Barlow Twins employs two identical networks, fed with two different versions of the batch samples, to construct two versions of the low-dimensional projections.
Afterwards it attempts to equate the cross-correlation matrix computed from the twin projections to identity, hence reducing the redundancy between different components of the projections.
This approach is relatable to several well-known objective functions for SSL, such as the information bottleneck objective~\cite{information-bottleneck}, or the \textsc{infoNCE} objective~\cite{info-nce}.

The idea of Barlow Twins has been ported recently to graph datasets by Graph Barlow Twins or G-BT~\cite{g-bt}.
Inspired by the image-augmentations proposed by Barlow Twins (cropping, color jittering, and blurring, etc.), G-BT adopts edge dropping and node feature masking to form the augmented views of the input graphs.
While this approach works for transductive node classification, its performance degrades for tasks involving link-prediction as demonstrated by the experiments in \secref{sec:experiments} because there is no explicit objective to preserve the information in links.

As we will see in \secref{sec:cov-los}, addition of this objective is non-trivial because it involves a careful modification of the original loss term from Barlow Twins.

\subsection{Graph Diffusion Convolution (GDC)} \label{sec:gdc}
GDC~\cite{diffusion} was proposed as a way to efficiently aggregate information from a large neighborhood.
This is achieved in two steps.
\begin{enumerate}
    \item \textbf{Diffusion: } First a dense diffusion matrix $\overline{\rmS}$ is constructed from the adjacency matrix using generalized graph diffusion as a denoising filter.
    \item \textbf{Sparsification: } This is the second step where either top $k$ entries of $\overline{\rmS}$ are selected in every row, or the entries below a threshold $\epsilon$ are set to $0$.
          We use $\rmS$ to denote the resulting sparse diffused matrix.
          The value of threshold can also be estimated from the intended average degree of the sparse graph \cite{diffusion}.
\end{enumerate}

$\rmS$ defines an alternate graph with weighted edges that carry more information than a binary adjacency matrix.
This sparse matrix, when used in place of the original adjacency matrix, improves graph learning for a variety of graph-based models such as degree corrected stochastic block model or DCSBM~\cite{DCSBM}, DeepWalk, GCN, GAT, GIN, and DGI, etc.

\section{Barlow Graph Auto-Encoder}
\begin{figure*}
    \centering
    \includegraphics[width=0.9\linewidth, trim={0cm 0cm 0cm 0cm},clip]{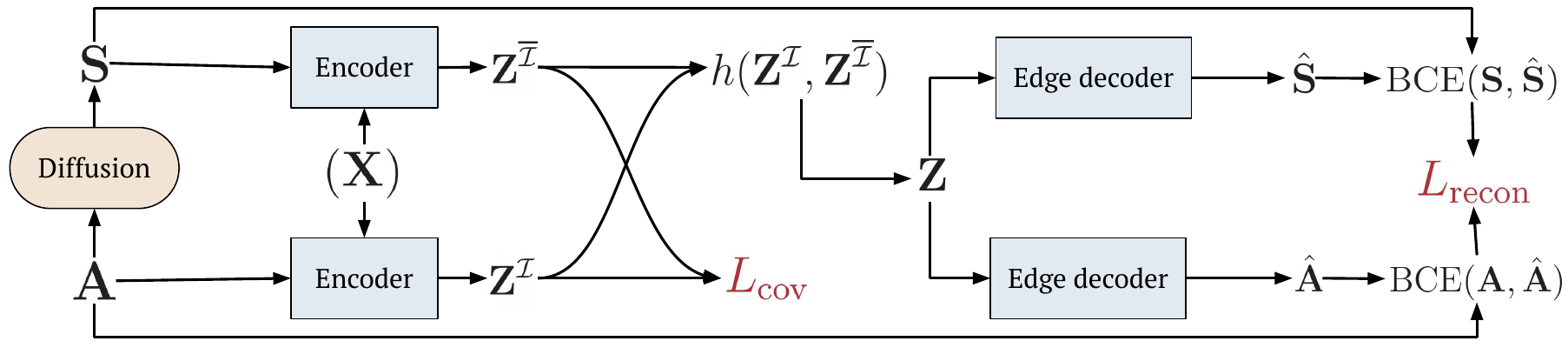}
    \caption{
        General model architecture.
        Based upon the input adjacency matrix $\rmA$, we get the diffusion matrix $\rmS$, which needs to be computed only once.
        The encoder yields the low dimensional projection matrices $\rmZ^{\gI}$ and $\rmZ^{\overline{\gI}}$ corresponding to $\gI$ and $\overline{\gI}$ respectively.
        Using a fusion function $h(\cdot)$, The projections $\rmZ^{\gI}$ and $\rmZ^{\overline{\gI}}$ are fused into a single projection $\rmZ$, which is then fed to the edge decoder to reconstruct $\Ahat$ and $\mathbf{\hat{S}}$.
    }
    \label{fig:model-architecture}
\end{figure*}

\subsection{Problem Formulation}
Suppose an undirected graph $\gG = (\gV, \gE)$ with the adjacency matrix $\rmA \in \{0,1\}^{N\times N}$ and optionally a matrix $\rmX \in \sR^{N \times F}$ of $F$-dimensional node features, $N$ being the number of nodes.
In addition, we construct a \textit{diffused} version of $\gG$ by building a diffusion matrix $\rmS \in \sR^{N \times N}$ from $\rmA$.
For brevity we use $\gI = (\rmA, \rmX)$ and $\overline{\gI} = (\rmS, \rmX)$ if features are available, otherwise $\gI = \rmA$ and $\overline{\gI} = \rmS$.
Given $d$ as the embedding-dimension size, we aim to optimize the model parameters for finding the network embeddings $\rmZ^{\gI} \in \sR^{N \times d}$ and $\rmZ^{\overline{\gI}} \in \sR^{N \times d}$ from $\gI$ and $\overline{\gI}$ such that:
\begin{enumerate}
    \item $\rmZ^{\gI}$ and $\rmZ^{\overline{\gI}}$ can be fused in a way that both $\rmA$ and $\rmS$ can be reconstructed from the fused embedding $\rmZ \in \sR^{N \times d}$.
          This allows the embedding to capture the local information from $\gI$ as well as the information in a larger neighborhood from $\overline{\gI}$.
    \item Same components of different projections have high covariance and vice versa for different components.
          This adds to the stability towards distortions and also reduces redundancy between different components of the embedding.
\end{enumerate}

Mathematically, the above objective is achieved by minimizing the following loss function
\begin{equation}
    L = L_{\mathrm{recon}} + \beta L_{\mathrm{cov}} \label{eq:general-loss}
\end{equation}
where $\beta$ is a hyperparameter of the algorithm to weigh between the reconstruction loss and the cross-covariance loss.
The general model architecture is given in \figref{fig:model-architecture}.
We first define the loss terms and then brief the modules of the architecture.

\subsection{Loss Terms}
\subsubsection{Reconstruction Loss ($L_{\mathrm{recon}}$)}
We aim to learn the model parameters in order to maximize the log probability of recovering both $\rmA$ and $\rmS$ from $\rmZ$.
This probability can be written as a marginalization over the joint distribution containing the latent variables $\rmZ$ as
\begin{align}
    \log\Big(p(\rmA, \rmS)\Big) & = \log\Big(\int p(\rmA, \rmS, \rmZ) d\rmZ\Big)                                               \\
                                & = \log\Big(\int p(\rmZ) p(\rmA | \rmZ) p(\rmS | \rmZ) d\rmZ \Big), \label{eq:log_prob_joint}
\end{align}
where the prior $p(\rmZ)$ is modelled as a unit gaussian.
\Eqref{eq:log_prob_joint} assumes conditional independence between $\rmA$ and $\rmS$ given $\rmZ$.

To ensure tractability, we introduce the approximate posterior $q(\rmZ | \gI, \overline{\gI})$ given by
\begin{align}
    q(\rmZ | \gI, \overline{\gI}) & = q(\rmZ^{\gI}\rmZ^{\overline{\gI}} | \gI, \overline{\gI}) \label{eq:q_complete}     \\
                                  & = q(\rmZ^{\gI} | \gI) q(\rmZ^{\overline{\gI}} | \overline{\gI}) \label{eq:q_broken},
\end{align}
where \eqref{eq:q_broken} follows from \eqref{eq:q_complete} because of the assumed conditional independence of $\rmZ^{\gI}$ and $\rmZ^{\overline{\gI}}$ given their respective inputs.
Both $q(\rmZ^{\gI} | \gI)$ and $q(\rmZ^{\overline{\gI}} | \overline{\gI})$ are modelled as Gaussians by a single encoder block that learns the parameters $\mu(\cdot)$ and $\sigma(\cdot)$ of the distribution conditioned on the given inputs $\gI$ and $\overline{\gI}$ respectively.
The term $L_{\mathrm{recon}}$ can now be considered as a negative of the ELBO bound derived as
\begin{alignat}{2}
    \log\big(p(\rmA, \rmS)\big) & \geq &  & - D_{KL}\Big(q(\rmZ^{\gI} | \gI) \ || \ \gN(\mathbf{0}, \rmI) \Big) \nonumber                               \\
                                &      &  & - D_{KL}\Big(q(\rmZ^{\overline{\gI}} | \overline{\gI}) \ || \ \gN(\mathbf{0}, \rmI) \Big) \nonumber         \\
                                &      &  & - \mathrm{BCE}\big(\Ahat, \rmA \big) - \mathrm{BCE}\big(\mathbf{\hat{S}}, \rmS \big) \label{eq:elbo_jensen} \\
                                & =    &  & \gL_{\mathrm{ELBO}}                                                                                         \\
                                & =    &  & - L_{\mathrm{recon}}, \label{eq:elbo_variational}
\end{alignat}
where $D_{KL}$ refers to the KL divergence, BCE is the binary cross-entropy, and the matrices $\Ahat$ and $\mathbf{\hat{S}}$ refer to reconstructed versions of $\rmA$ and $\rmS$ respectively.
The inequality in \ref{eq:elbo_jensen} follows from Jensen's inequality.
It is worth noticing that BCE is computed based upon the edges constructed from the fused embedding $\rmZ$.
For detailed derivation, we refer the reader to the supplementary material.

The reconstruction loss in \ref{eq:elbo_variational} refers to the variational variant \vours.
For the non-variational case of \ours, the KL divergence terms get dropped, leaving only the two BCE terms.

\subsubsection{Covariance Loss ($L_{\mathrm{cov}}$)} \label{sec:cov-los}
The correlation-loss in \cite{barlow} (also employed in \cite{g-bt}) involves normalization by the standard deviation of the embedding vectors, centered across the input batch (equation $2$ in \cite{barlow}).
While this works for images and for nodes of the graphs, it has a tendency to obscure the information in the relative strengths of the links in a graph.
For graph-datasets, we often replace cosine similarity with dot products, followed by a sigmoid (as done in ~\cite{vgae}, \cite{evgae},and \cite{mvgrl}, etc.) as it helps in preventing the information in the magnitude of the vectors.
Following this approach, instead of computing the cross-correlation matrix, we compute the cross-covariance matrix and use the $\mathrm{sigmoid}$ function to individually normalize the absolute entries $c_{\ell m}$.
The loss $L_{\mathrm{cov}}$ is then computed as a summation of two terms corresponding to the mean of diagonal elements and off-diagonal elements of $\gC$.
\begin{align}
    L_{\mathrm{cov}} = & - \frac{1}{N}\sum \limits_{\ell = m} \mathrm{log}(c_{mm}) \nonumber                                 \\
                       & - \frac{\lambda}{N(N - 1)}\sum \limits_{\ell \neq m} \mathrm{log}(1 - c_{\ell m}), \label{eq:l_cov}
\end{align}
where $\lambda$ defines the trade-off between the two terms.
The first term of \eqref{eq:l_cov} is the invariance term.
When minimized, it makes the embedding stable towards distortions.
The second term refers to the cross-covariance between different components of $\gC$.
When minimized, it reduces the redundancy between different components of the vectors.
The entries $c_{\ell m}$ of $\gC$ are given by
\begin{align}
    c_{\ell m} = \mathrm{sigmoid}\Big(\big| \sum\limits_{b=1}^{|\gB|} (z_{b\ell}^{\gI} - \overline{z}_{\ell}^{\gI}) (z_{bm}^{\overline{\gI}} - \overline{z}_{m}^{\overline{\gI}})\big|\Big),
\end{align}
where $b$ indexes the batch $\gB$ with size $|\gB|$ and the $\ell$-th component of the latent embedding $\rvz^{\gI}_b$ is denoted by $z_{b\ell}^{\gI}$.
The empirical means across embeddings $\rmZ^{\gI}$ and $\rmZ^{\overline{\gI}}$ are denoted by $\overline{z}^{\gI}$ and $\overline{z}^{\overline{\gI}}$ respectively.
It is also worth noticing that the underlying objective is the same for \eqref{eq:l_cov} as well as the original correlation-based loss in Barlow Twins i.e., $\gC$ should be as close to identity matrix as possible.

\subsection{Model Architecture Blocks}
We now describe the modules leading to the loss terms in \eqref{eq:general-loss} as shown in \figref{fig:model-architecture}.
\subsubsection{Diffusion:}
The generalized diffusion to construct $\rmS$ from $\rmA$ is given by
\begin{equation}
    \rmS = \sum \limits_{k = 0}^{\infty} \theta_k \rmT^k, \label{eq:general-diffusion}
\end{equation}
where $\rmT$ is the generalized transition matrix and $\theta_k$ are the weighting coefficients.
There can be multiple possibilities for $\theta_k$ and $\rmT^k$ while ensuring the convergence of \eqref{eq:general-diffusion} such as the ones proposed in~\cite{diffusion-possibilities-2},~\cite{diffusion-possibilities-1}, and~\cite{diffusion} etc.
In this work, we report the case of Personalized PageRank (PPR:~\cite{pagerank}) as it consistently gives better results with \ours/\vours.
For the detailed results including the Heat Kernel \cite{diffusion}, we refer the reader to the supplementary material.

PPR kernel corresponds to $\rmT = \rmA \rmD^{-1}$ and $\theta_k = \alpha(1 - \alpha)^k$, where $\rmD$ is the degree matrix and $\alpha \in (0, 1)$ is the teleport probability.
The corresponding symmetric transition matrix is given by $\rmT = \rmD^{-1/2} \rmA \rmD^{-1/2}$.
Substitution of $\rmT$ and $\theta_k$ into \eqref{eq:general-diffusion} leads to a closed form solution for diffusion using PPR kernel as
\begin{equation}
    \rmS = \alpha \Big(\rmI - (1 - \alpha)\rmD^{-1/2} \rmA \rmD^{-1/2} \Big)^{-1}. \label{eq:ppr-diffusion}
\end{equation}
\Eqref{eq:ppr-diffusion} restricts the use of PPR-based diffusion for large graphs.
However, in practice, there exist multiple approaches to efficiently approximate \eqref{eq:ppr-diffusion} such as the ones proposed by~\cite{ppr_approx_1} and~\cite{ppr_approx_2}.
In all the results reported in this paper, we use the approximate version of \eqref{eq:ppr-diffusion}.
Diffusion works well for graphs with high homophily.
So \ours/\vourss also target similar networks.

\subsubsection{Encoder:}
This module is responsible for projecting the information in $\gI$ and $\overline{\gI}$ into $d$-dimensional embeddings $\rmZ^{\gI}$ and $\rmZ^{\overline{\gI}}$ respectively.
Following~\cite{barlow}, we use a single encoder to encode both versions of the input graph.
Our framework is general in the sense that any reasonable encoder can be plugged in to get the learnable projections.

In our work, we have considered two options for the encoder block, consequently leading to two variants of the framework.
\begin{itemize}
    \item For \ours, we use a single layer GCN encoder.
    \item For \vours, we employ a simple variational encoder that learns the parameters $\mu(\cdot)$ and $\sigma(\cdot)$ of a gaussian distribution conditioned upon the input samples.
          The latent samples can then be generated by following the reparameterization trick~\cite{vae}.
\end{itemize}

\subsubsection{Fusion Function:}
The function $h(\cdot)$ is used to fuse $\rmZ^{\gI}$ and $\rmZ^{\overline{\gI}}$ into a single matrix $\rmZ$.
In this work we define $\rmZ$ as a weighted sum of $\rmZ^{\gI}$ and $\rmZ^{\overline{\gI}}$ as
\begin{equation}
    h(\rvz_i^{\gI}, \rvz_i^{\overline{\gI}}) = \phi_{i}^{\gI} \rvz_i^{\gI} + \phi_{i}^{\overline{\gI}} \rvz_i^{\overline{\gI}},
\end{equation}
where the weights $\{\phi_{i}^{\gI}\}_{i = 1}^{N}$ and $\{\phi_{i}^{\overline{\gI}}\}_{i = 1}^{N}$ can either be fixed or learned.
In this work we report two variants of the fusion function.
\begin{itemize}
    \item Fixing $\phi_{i}^{\gI} = \phi_{i}^{\overline{\gI}} = 0.5 \ \forall i$.
    \item Learning $\{\phi_{i}^{\gI}\}_{i = 1}^{N}$ and $\{\phi_{i}^{\overline{\gI}}\}_{i = 1}^{N}$ using attention mechanism.
\end{itemize}

For attention, we compute the dot products of different embeddings of the same node with respective learnable weight vectors, followed by LeakyReLU activation.
Afterwards a $\mathrm{softmax}$ is applied to get the probabilistic weight assignments, i.e.
\begin{align}
    \phi_{i}^{\gI}            & = \frac{\exp\Big(\mathrm{A}\big(\rvw_1^T\rvz_i^{\gI} \big)\Big)}{\exp\Big(\mathrm{A}\big(\rvw_1^T \rvz_i^{\gI}\big)\Big) + \exp\Big(\mathrm{A}\big(\rvw_2^T \rvz_i^{\overline{\gI}} \big)\Big)}, \label{eq:fusion-with-attention-phi1i} \\
    \phi_{i}^{\overline{\gI}} & = 1 - \phi_{i}^{\gI}. \label{eq:fusion-with-attention-phi2i}
\end{align}

where $\textrm{A}(\cdot)$ is the LeakyReLU activation function and $\rvw_1, \rvw_2 \in \sR^{d}$ are learnable weight vectors.

\subsubsection{Decoder}
Since our approach is auto-encoder-based, we use the edge decoder as proposed by~\cite{vgae}, to reconstruct the entries $\hat{a}_{ij}$ of $\Ahat$ as
\begin{align}
    \hat{a}_{ij} = \sigmoid(\rvz_i^T\rvz_j)
\end{align}
The entries of $\mathbf{\hat{S}}$ can also be reconstructed in similar fashion.

\section{Experiments} \label{sec:experiments}
This section describes the datasets and the experiments conducted to evaluate the efficacy of our approach.
We choose eight benchmark datasets including Wikipedia articles (WikiCS)~\cite{dataset:wikics}, Amazon co-purchase data networks (AmazonPhoto and AmazonComputers~\cite{dataset:amzCompAndPhotos}), extracts from Microsoft Academic Graph (CoauthorCS and CoauthorPhysics)~\cite{dataset:coauthorCsAndPhysics}, and citation networks (Cora, CiteSeer and PubMed)~\cite{dataset:citation}.
The basic characteristics of these datasets are briefed in \tabref{tab:datasets}.
We first report the results for link prediction.
The network embedding learned by \ourss and \vourss is unsupervised as no node labels are used during training.
Hence, to measure the quality of the network embedding, we analyze our approach for two downstream tasks: clustering and transductive node classification.
For the interested readers, the supplementary material contains a detailed analysis of \ourss and \vourss in different settings.
We use the AWS EC2 instance type \texttt{g4dn.4xlarge} with $16$GB GPU for training.
For reproducibility, the implementation details of all the experiments along with the code are provided in \cite{code}.
For all the experiments, we report publicly available results from our competitors.
\begin{table}[]
    \caption{
        Datasets used for evaluation.
    }
    \centering
    \resizebox{1\linewidth}{!}{%
        {\setlength{\tabcolsep}{0.5em}%
                \begin{tabular}{lrrrr}
                    \toprule
                    \textbf{Dataset}                                             & \textbf{Nodes} & \textbf{Edges} & \textbf{Features} & \textbf{Classes} \\
                    \midrule
                    \textbf{Cora~\cite{dataset:citation}}                        & 2,708          & 5,297          & 1,433             & 7                \\
                    \textbf{CiteSeer~\cite{dataset:citation}}                    & 3,312          & 4,732          & 3,703             & 6                \\
                    \textbf{PubMed~\cite{dataset:citation}}                      & 19,717         & 44,338         & 500               & 3                \\
                    \textbf{WikiCS~\cite{dataset:wikics}}                        & 11,701         & 216,123        & 300               & 10               \\
                    \textbf{AmazonComputers~\cite{dataset:amzCompAndPhotos}}     & 13,752         & 245,861        & 767               & 10               \\
                    \textbf{AmazonPhotos~\cite{dataset:amzCompAndPhotos}}        & 7,650          & 119,081        & 745               & 8                \\
                    \textbf{CoauthorCS~\cite{dataset:coauthorCsAndPhysics}}      & 18,333         & 81,894         & 6,805             & 15               \\
                    \textbf{CoauthorPhysics~\cite{dataset:coauthorCsAndPhysics}} & 34,493         & 247,962        & 8,415             & 5                \\
                    \bottomrule
                \end{tabular}
            }
    }\label{tab:datasets}
\end{table}

\subsection{Link Prediction}
\begin{table*}[tbh]
    \caption{
        Link prediction performance, as evaluated by AUC and AP metrics.
        The best results are styled as bold and second best are underlined.
    }
    \centering
    \resizebox{0.9\linewidth}{!}{%
        {\setlength{\tabcolsep}{1.2em}%
                \begin{tabular}{lrrrrrrrrrr}
                    \toprule
                    \textbf{Algorithm} & \multicolumn{2}{c}{\textbf{Cora}} & \multicolumn{2}{c}{\textbf{CiteSeer}} & \multicolumn{2}{c}{\textbf{PubMed}} & \multicolumn{2}{c}{\textbf{CoauthorCS}} & \multicolumn{2}{c}{\textbf{AmazonPhoto}}                                                                                                                                                                             \\
                    \cmidrule(l{4pt}r{4pt}){2-3} \cmidrule(l{4pt}r{4pt}){4-5} \cmidrule(l{4pt}r{4pt}){6-7} \cmidrule(l{4pt}r{4pt}){8-9} \cmidrule(l{4pt}r{4pt}){10-11}
                                       & \multicolumn{1}{l}{\textbf{AUC}}  & \multicolumn{1}{l}{\textbf{AP}}       & \multicolumn{1}{l}{\textbf{AUC}}    & \multicolumn{1}{l}{\textbf{AP}}         & \multicolumn{1}{l}{\textbf{AUC}}         & \multicolumn{1}{l}{\textbf{AP}} & \multicolumn{1}{l}{\textbf{AUC}} & \multicolumn{1}{l}{\textbf{AP}} & \multicolumn{1}{l}{\textbf{AUC}} & \multicolumn{1}{l}{\textbf{AP}} \\
                    \midrule
                    DeepWalk           & 83.10                             & 85.00                                 & 80.50                               & 83.60                                   & 84.40                                    & 84.10                           & 91.74                            & 91.19                           & 91.48                            & 90.79                           \\
                    \midrule
                    GAE                & 91.00                             & 92.00                                 & 89.50                               & 89.90                                   & 96.40                                    & 96.50                           & 94.09                            & 93.86                           & 93.86                            & 92.96                           \\
                    VGAE               & 91.40                             & 92.60                                 & 90.80                               & 92.00                                   & 94.40                                    & 94.70                           & 89.60                            & 89.36                           & 92.05                            & 92.02                           \\
                    ARGA               & 92.40                             & 93.20                                 & 91.90                               & 93.00                                   & 96.80                                    & 97.10                           & 91.99                            & 92.54                           & \textbf{96.10}                   & \textbf{95.40}                  \\
                    ARVGA              & 92.40                             & 92.60                                 & 92.40                               & 93.00                                   & 96.50                                    & 96.80                           & 93.32                            & 93.32                           & 92.70                            & 90.90                           \\
                    GALA               & 92.10                             & 92.20                                 & 94.40                               & 94.80                                   & 91.50                                    & 89.70                           & 93.81                            & 94.49                           & 91.80                            & 91.00                           \\
                    \midrule
                    DGI                & 89.80                             & 89.70                                 & 95.50                               & 95.70                                   & 91.20                                    & 92.20                           & 94.87                            & 94.34                           & 92.24                            & 92.14                           \\
                    GIC                & 93.50                             & 93.30                                 & 97.00                               & 96.80                                   & 93.70                                    & 93.50                           & 95.03                            & 94.94                           & 92.70                            & 92.34                           \\
                    GMI                & 95.10                             & 95.60                                 & 97.80                               & 97.40                                   & 96.37                                    & 96.04                           & \underline{96.37}                & 95.04                           & 93.88                            & 92.67                           \\
                    GCA                & 95.75                             & 95.47                                 & 96.44                               & 96.49                                   & 95.28                                    & 95.52                           & 96.31                            & \textbf{96.28}                  & 93.25                            & 92.74                           \\
                    MVGRL              & 90.52                             & 90.45                                 & 92.89                               & 92.89                                   & 92.45                                    & 92.17                           & 95.17                            & 95.58                           & 92.89                            & 92.45                           \\
                    \midrule
                    G-BT               & 87.46                             & 86.84                                 & 93.42                               & 93.01                                   & 94.53                                    & 94.26                           & 92.64                            & 91.40                           & 95.12                            & 95.45                           \\
                    \midrule
                    BGAE               & \underline{98.52}                 & \underline{98.42}                     & \underline{98.59}                   & \textbf{98.61}                          & 97.78                                    & 97.68                           & 96.31                            & 95.44                           & 95.01                            & 94.24                           \\
                    BGAE + Att         & \textbf{98.79}                    & \textbf{98.73}                        & 98.56                               & \underline{98.57}                       & \textbf{98.06}                           & \textbf{98.03}                  & \textbf{96.51}                   & \underline{95.65}               & \underline{95.18}                & \underline{94.49}               \\
                    BVGAE              & 97.87                             & 97.62                                 & \textbf{98.63}                      & \underline{98.57}                       & \underline{97.93}                        & \underline{97.89}               & 96.12                            & 95.13                           & 94.61                            & 94.29                           \\
                    BVGAE + Att        & 98.03                             & 97.77                                 & 98.23                               & 98.08                                   & 97.77                                    & 97.74                           & 96.21                            & 95.34                           & 94.97                            & 94.28                           \\
                    \bottomrule
                \end{tabular}
            }
    }\label{tab:link-prediction}
\end{table*}

\subsubsection{Comparison Methods} \label{sec:exp:link-pred:baselines}
For link prediction, we select 12 competitors.
We start with \textbf{DeepWalk}~\cite{deepwalk} as the baseline.

\textbf{Auto-encoder based Architectures}: We include graph auto-encoder or \textbf{GAE} which aims to reconstruct the adjacency matrix for the input graph.
The variational graph auto-encoder or \textbf{VGAE}~\cite{vgae} is its variational counterpart that extends the idea of variational auto-encoder or VAE~\cite{vae} to graph domain.
In the case of adversarially regularized graph auto-encoder or \textbf{ARGA}~\cite{arvga}, the latent representation is forced to match the prior via an adversarial training scheme.
Just like VGAE, there exists a variational alternative to ARGA, known as adversarially regularized variational graph auto-encoder or \textbf{ARVGA}.
\textbf{GALA}~\cite{gala} learns network embedding by treating encoding and decoding steps as Laplacian smoothing and Laplace sharpening respectively.

\textbf{Contrastive Methods}: \textbf{DGI}~\cite{dgi} leverages Deep-Infomax~\cite{deep-infomax} for graph datasets.
Graph InfoClust or \textbf{GIC}~\cite{gic} learns network embedding by maximizing the mutual information with respect to the graph-level summary as well as the cluster-level summaries.
\textbf{GMI}~\cite{gmi} aims to learn node representations while aiming to improve generalization performance via added contrastive regularization.
\textbf{GCA}~\cite{gca} proposes adaptive augmentation techniques to contrast views between nodes and subgraphs or structurally transformed graphs.
\textbf{MVGRL}~\cite{mvgrl} learns graph embedding by contrasting multiple views of the input data.

In addition, we include \textbf{G-BT} \cite{g-bt} which is another approach making use of the redundancy-minimization principle introduced in \cite{barlow} as discussed in \secref{sec:related:barlow}.

For link prediction, we skip the three datasets where many public results are missing and report these partial results in the supplementary material.

\subsubsection{Settings} \label{sec:exp:link-pred:settings}
For link prediction, we follow the same link split as adopted by our competitors, i.e., we split the edges into the training, validation, and test sets containing 85\%, 5\%, and 10\% links respectively.
For all the competitors we keep the same settings as given by the authors.
For the methods where multiple variants are given by the authors(e.g., ARGA, ARVGA, etc.), we report the best results amongst all the variants.
For our approach, we keep the latent dimension to $512$ for all the experiments except CoauthorPhysics where $d = 128$ to avoid out-of-memory issues.
We use a single layer GCN encoder for \ours, and two GCN encoders to output the parameters $\mu(\cdot)$ and $\sigma(\cdot)$ in case of the variational encoder block of \vours.
For all the experiments, we get $\rmS$ by setting the average degree to $25$.
The value of the hyperparameter $\lambda$ in \eqref{eq:l_cov} is set to $5e^{-3}$ for all the experiments.
This is the same as proposed in Barlow-Twins\cite{barlow}.
Adam~\cite{adam} is used as the optimizer with learning rate and rate decay set to 0.01 and $5e^{-6}$.
The hyperparameter $\beta$ is fixed to $1$ for all experiments.
Instead of computing a closed-form solution, it is sufficient to compute the BCE loss using the samples $\rvz$.
For this, we follow other auto-encoder-based approaches such as~\cite{vgae,evgae} for dataset splits and sampling of positive/negative edges for every training iteration.
For evaluation, we report the area under the curve (AUC) and average precision (AP) metrics.
All the results are the average of 10 runs.
Further implementation details can be found in~\cite{code}.

\begin{table*}[tbh]
    \captionsetup{width=1\linewidth}
    \caption{
        Transductive node classification performance, as evaluated by accuracy.
        The best results are styled as bold and second best are underlined.
        OOM refers to Out-of-Memory.
    }
    \centering
    \resizebox{1\linewidth}{!}{%
        {\setlength{\tabcolsep}{1em}%
                \begin{tabular}{lrrrrrrrr}
                    \toprule
                    \textbf{Algorithm} & \multicolumn{1}{c}{\textbf{Cora}} & \multicolumn{1}{c}{\textbf{CiteSeer}} & \multicolumn{1}{c}{\textbf{PubMed}} & \multicolumn{1}{c}{\textbf{WikiCS}} & \multicolumn{1}{c}{\textbf{CoauthorCS}} & \multicolumn{1}{c}{\textbf{CoauthorPhysics}} & \multicolumn{1}{c}{\textbf{AmazonComputers}} & \multicolumn{1}{c}{\textbf{AmazonPhoto}} \\
                    \midrule
                    Raw                & 47.87                             & 49.33                                 & 69.11                               & 71.98                               & 90.37                                   & 93.58                                        & 73.81                                        & 78.53                                    \\
                    DeepWalk           & 70.66                             & 51.39                                 & 74.31                               & 77.21                               & 87.70                                   & 94.90                                        & 86.28                                        & 90.05                                    \\
                    \midrule
                    GAE                & 71.53                             & 65.77                                 & 72.14                               & 70.15                               & 90.01                                   & 94.92                                        & 85.18                                        & 91.68                                    \\
                    VGAE               & 75.24                             & 69.05                                 & 75.29                               & 75.63                               & 92.11                                   & 94.52                                        & 86.44                                        & 92.24                                    \\
                    ARGA               & 74.14                             & 64.14                                 & 74.12                               & 66.88                               & 89.41                                   & 93.10                                        & 84.39                                        & \underline{92.68}                        \\
                    ARVGA              & 74.38                             & 64.24                                 & 74.69                               & 67.37                               & 88.54                                   & 94.30                                        & 84.66                                        & 92.49                                    \\
                    \midrule
                    DGI                & 81.68                             & 71.47                                 & 77.27                               & 75.35                               & 92.15                                   & 94.51                                        & 83.95                                        & 91.61                                    \\
                    GIC                & 81.73                             & 71.93                                 & 77.33                               & 77.28                               & 89.40                                   & 93.10                                        & 84.89                                        & 92.11                                    \\
                    GRACE              & 80.04                             & 71.68                                 & 79.53                               & \textbf{80.14}                      & 92.51                                   & 94.70                                        & 87.46                                        & 92.15                                    \\
                    GMI                & 83.05                             & \underline{73.03}                     & 80.10                               & 74.85                               & OOM                                     & OOM                                          & 82.21                                        & 90.68                                    \\
                    GCA                & 82.10                             & 71.30                                 & 80.20                               & 78.23                               & 92.95                                   & \textbf{95.73}                               & 88.94                                        & 92.53                                    \\
                    MVGRL              & 82.90                             & 72.60                                 & 79.40                               & 77.52                               & 92.11                                   & 92.11                                        & 87.52                                        & 91.74                                    \\
                    BGRL               & 82.70                             & 71.10                                 & 79.60                               & \underline{79.98}                   & 93.31                                   & 95.56                                        & 89.68                                        & \textbf{92.87}                           \\
                    \midrule
                    G-BT               & 80.80                             & 73.00                                 & 80.00                               & 76.65                               & 92.95                                   & 95.07                                        & 88.14                                        & 92.63                                    \\
                    \midrule
                    BGAE               & \underline{83.51}                 & 72.43                                 & \textbf{81.84}                      & 78.93                               & \textbf{93.76}                          & 95.01                                        & \underline{92.24}                            & 91.10                                    \\
                    BGAE + Att         & \textbf{83.60}                    & 72.41                                 & \underline{80.95}                   & 79.53                               & \textbf{93.76}                          & \underline{95.64}                            & \textbf{92.44}                               & 91.89                                    \\
                    BVGAE              & 82.62                             & 72.97                                 & 80.02                               & 77.52                               & 93.25                                   & 95.13                                        & 89.19                                        & 89.38                                    \\
                    BVGAE + Att        & 82.57                             & \textbf{73.09}                        & 80.25                               & 77.82                               & 93.15                                   & 95.60                                        & 89.91                                        & 89.98                                    \\
                    \bottomrule
                \end{tabular}
            }
    }\label{tab:classification}
\end{table*}

\subsubsection{Results}
\Tabref{tab:link-prediction} gives the results for the link prediction task.
For our approach, we give the results for \ourss as well as \vours, both with and without attention.
We can observe that our approach achieves the best or second best results in all the datasets.
Overall the variant of \ourss with attention performs well across all the datasets and metrics.
This validates the choice of attention as a fusion function.
Among the competitors, the contrastive approaches perform relatively better across all the datasets.
One exception is AmazonPhoto where ARGA achieves the best results.
However, as the next sections demonstrate, the performance of ARGA/ARVGA degrades for downstream node classification and clustering.
Another thing to note is the results of G-BT.
As the training epochs go on for G-BT, the results degrade rapidly, often by about 30\% of the results reported in \tabref{tab:link-prediction}.
Apart from AmazonPhotos, there is a healthy margin between \ourss and G-BT mainly because unlike \ours, G-BT does not explicitly preserve the information in the links.

\subsection{Transductive Node Classification}
\subsubsection{Comparison Methods}
We compare with 14 competitors for transductive node classification, using raw features and DeepWalk(with features) as the baselines.
In addition to the methods briefed in \secref{sec:exp:link-pred:baselines}, we include two more contrastive methods i.e. GRACE and BGRL:
\textbf{GRACE}~\cite{grace} learns network embedding by making use of multiple views and contrasting the representation of a node with its raw information (e.g., node features) or neighbors' representations in different views.
\textbf{BGRL}~\cite{bgrl} eliminates the need for negative samples by minimizing invariance between two augmented versions of mini-batches of graphs.

\subsubsection{Settings}\label{sec:exp:classification:settings}
The training phase uses the same settings as reported in \secref{sec:exp:link-pred:settings}.
For transductive node classification, we do not need to split the edges into training/validation/test sets.
So we use all the edges for self-supervised learning of the node embeddings.
For evaluating the embedding, a logistic regression head is used with \texttt{lbfgs} solver.
For this, we use the default settings of the \texttt{scikit-learn} package.
For citation datasets (Cora, CiteSeer, and PubMed), we follow the standard public splits for training/validation/test sets used in many previous works such as~\cite{classification-split,dgi,j-enc,gcn}, i.e., 20 labels per class for training, 500 samples for validation, and 1000 for testing.
For WikiCS, we average over the 20 splits that are publicly provided.
For the rest of the datasets (AmazonPhoto, AmazonComputers, CoauthorPhysics, and coauthorCS), we follow the split configuration of B-JT, i.e. generate random splits with training, validation, and test sets containing 10\%, 10\%, and 80\% nodes respectively.
For evaluation, we use accuracy as the metric.

\subsubsection{Results}
\Tabref{tab:classification} gives the comparison between different algorithms for transductive node classification.
We can again observe consistently good results by our approach for all eight datasets.
For this task, the margin is rather small, especially for Cora and CiteSeer, compared to the best competitor i.e. GMI.
Nonetheless, our point still stands well-conveyed that our approach performs on par with the well-known network embedding techniques for transductive node classification.
A comparison of \tabref{tab:classification} with \tabref{tab:link-prediction} demonstrates inconsistencies in the performance of our competitors for the two tasks.
This is mainly because either the competitors do not explicitly preserve information in the links (e.g. MVGRL, G-BT, etc), or link prediction is their main focus (e.g., in GAE/ARGA, etc).
For instance, ARGA performed reasonably well for link prediction, but fails to give a similar consistent performance across all datasets in \tabref{tab:classification}.
On the other hand, MVGRL performs well in \tabref{tab:classification}, although its performance suffered in \tabref{tab:link-prediction}.

\subsection{Node Clustering}
\begin{table*}[tbh]
    \captionsetup{width=1\linewidth}
    \caption{
        Node clustering performance, as evaluated by NMI.
        The best results are styled as bold and second best are underlined.
        OOM refers to Out-of-Memory.
    }
    \centering
    \resizebox{0.95\linewidth}{!}{%
        {\setlength{\tabcolsep}{0.7em}%
                \begin{tabular}{lrrrrrrrr}
                    \toprule
                    \textbf{Algorithm} & \multicolumn{1}{c}{\textbf{Cora}} & \multicolumn{1}{c}{\textbf{CiteSeer}} & \multicolumn{1}{c}{\textbf{PubMed}} & \multicolumn{1}{c}{\textbf{WikiCS}} & \multicolumn{1}{c}{\textbf{CoauthorCS}} & \multicolumn{1}{c}{\textbf{CoauthorPhysics}} & \multicolumn{1}{c}{\textbf{AmazonComputers}} & \multicolumn{1}{c}{\textbf{Amazon-Photos}} \\
                    \midrule
                    K-means            & 32.10                             & 30.50                                 & 0.10                                & 18.20                               & 64.20                                   & 48.90                                        & 16.60                                        & 28.20                                      \\
                    \midrule
                    GAE                & 42.90                             & 17.60                                 & 27.70                               & 24.30                               & 73.10                                   & 54.50                                        & 44.10                                        & 61.60                                      \\
                    VGAE               & 43.60                             & 15.60                                 & 22.90                               & 26.10                               & 73.30                                   & 56.30                                        & 42.30                                        & 53.00                                      \\
                    ARGA               & 44.90                             & 35.00                                 & 30.50                               & 27.50                               & 66.80                                   & 51.20                                        & 23.50                                        & 57.70                                      \\
                    ARVGA              & 52.60                             & 33.80                                 & 29.00                               & 28.70                               & 61.60                                   & 52.60                                        & 23.70                                        & 45.50                                      \\
                    \midrule
                    DGI                & 41.10                             & 31.50                                 & 27.70                               & 31.00                               & 74.70                                   & 67.00                                        & 31.80                                        & 37.60                                      \\
                    GRACE              & 46.18                             & 38.29                                 & 16.27                               & 42.82                               & 75.62                                   & OOM                                          & 47.93                                        & 65.13                                      \\
                    GCA                & 55.70                             & 37.40                                 & 28.90                               & 29.90                               & 73.50                                   & 59.40                                        & 42.60                                        & 34.40                                      \\
                    MVGRL              & 60.90                             & \textbf{44.00}                        & 31.50                               & 26.30                               & 74.00                                   & 59.40                                        & 24.40                                        & 34.40                                      \\
                    \midrule
                    G-BT               & 43.40                             & 41.57                                 & 29.52                               & 27.46                               & 74.37                                   & 59.8                                         & 65.55                                        & 52.39                                      \\
                    \midrule
                    BGAE               & \textbf{62.42}                    & 43.36                                 & \underline{38.46}                   & \underline{45.80}                   & \underline{80.10}                       & \underline{68.01}                            & \textbf{66.98}                               & \underline{67.13}                          \\
                    BGAE + Att         & \underline{62.27}                 & \underline{43.84}                     & \textbf{38.59}                      & \textbf{46.93}                      & \textbf{80.30}                          & \textbf{68.12}                               & \underline{66.93}                            & \textbf{67.43}                             \\
                    BVGAE              & 59.60                             & 43.29                                 & 37.41                               & 40.78                               & 79.01                                   & 67.10                                        & 60.98                                        & 61.33                                      \\
                    BVGAE + Att        & 59.82                             & 43.27                                 & 37.47                               & 40.86                               & 79.42                                   & 67.06                                        & 61.44                                        & 61.62                                      \\
                    \bottomrule
                \end{tabular}
            }
    }\label{tab:clustering}
\end{table*}

\subsubsection{Settings} \label{sec:exp:clustering:settings}
For clustering, we choose 10 methods in total for comparison, with the baseline established by K-Means.
The experimental configuration for node clustering follows the same pattern as in \secref{sec:exp:link-pred:settings}.
For clustering, we use all the edges just like in \secref{sec:exp:classification:settings}, i.e., all the edges are used for self-supervised learning of the node-embeddings.
Afterward, we use K-Means to infer the cluster assignments from the embeddings.
For evaluation, we use normalized mutual information (NMI) as the metric.
\subsubsection{Results}
The results of the experiments for downstream node clustering are given in \tabref{tab:clustering}.
Here again, we perform consistently well for all the datasets except CiteSeer, where we achieve the second-best results by a small margin.
It is worth noticing that apart from CiteSeer, we achieve both the best and the second results using different variants of \ours.
A comparison of \tabref{tab:clustering} with \tabref{tab:link-prediction} and \tabref{tab:classification} again highlights that no competitor algorithm performs consistently well for all the tasks and datasets.
For instance, ARGA performed well on some datasets in \tabref{tab:link-prediction}.
However, its performance suffers in \tabref{tab:clustering}.
Similarly, GMI, which performs well in \tabref{tab:classification}, is outperformed by many other algorithms in node clustering.
On the other hand, the algorithms such as GIC, that perform well in \tabref{tab:clustering} are outperformed by others in \tabref{tab:classification}.
This highlights the task-specific nature of the network embedding learned by different competitors and also shows the efficacy of \ourss across multiple tasks and datasets.

\section{Conclusion}
This work proposes a simple yet effective auto-encoder based approach for network embedding that simultaneously employs the information in the immediate and larger neighborhoods.
To construct a uniform network embedding, the two information sources are efficiently coupled using the redundancy-minimization principle.
We propose two variants, \ourss and \vours, depending upon the type of encoder block.
To construct larger neighborhood from the immediate neighborhood, we use graph-diffusion.
Our work is restricted to the networks with high homophily, because diffusion only works well for such networks.
As demonstrated by the extensive experimentation, our approach is on par with the well-known baselines, often outperforming them over a variety of tasks such as link prediction, clustering, and transductive node classification.

\clearpage
\bibliographystyle{splncs04}
\bibliography{references}

\clearpage
\setcounter{page}{1}
\section*{\LARGE{SUPPLEMENTARY MATERIAL}}
Throughout the supplementary material, we follow the notation and references from the main paper.
\section{DERIVATION OF $L_{\mathrm{recon}}$ FOR VARIATIONAL CASE}
As mentioned in the paper, we aim to learn the model parameters $\theta$ to maximize the log probability of recovering the joint probability of $\rmA$ and $\rmS$ from $\rmZ$, given as
\begin{align}
    \log\Big(p(\rmA, \rmS)\Big) & = \log\Big(\int p(\rmA, \rmS, \rmZ) d\rmZ\Big)                     \\
                                & = \log\Big(\int p(\rmZ) p(\rmA | \rmZ) p(\rmS | \rmZ) d\rmZ \Big).
\end{align}
Here we assume conditional independence between $\rmA$ and $\rmS$ given $\rmZ$.
The approximate posterior, introduced for tractability, is given as
\begin{align}
    q(\rmZ | \gI, \overline{\gI}) & = q(\rmZ^{\gI}\rmZ^{\overline{\gI}} | \gI, \overline{\gI})                                   \\
                                  & = q(\rmZ^{\gI} | \gI) q(\rmZ^{\overline{\gI}} | \gI, \rmZ^{\gI})  \label{eq:q_complete_supp} \\
                                  & = q(\rmZ^{\gI} | \gI) q(\rmZ^{\overline{\gI}} | \overline{\gI}) \label{eq:q_broken_supp},
\end{align}
where \eqref{eq:q_broken_supp} follows from \eqref{eq:q_complete_supp} because of the assumed conditional independence of $\rmZ^{\gI}$ and $\rmZ^{\overline{\gI}}$ given their respective inputs $\gI$ and $\overline{\gI}$.
The corresponding prior $p(\rmZ)$ is assumed as a joint of i.i.d. Gaussians, i.e.
\begin{align}
    p(\rmZ) = p(\rmZ^{\gI})p(\rmZ^{\overline{\gI}}) = \gN(\mathbf{0}, \rmI) \gN(\mathbf{0}, \rmI). \label{eq:p_broken}
\end{align}
So $L_{\mathrm{recon}}$ can now be considered as a negative of the ELBO bound derived as
    {\fontsize{8pt}{10pt}
        \begin{alignat}{2}
            \log\big(p(\rmA, \rmS)\big) & =    &  & \log\Big(\int p(\rmZ) p(\rmA | \rmZ) p(\rmS | \rmZ) d\rmZ \Big)                                                                              \\
                                        & =    &  & \log\Big(\int \frac{p(\rmZ) p(\rmA | \rmZ) p(\rmS | \rmZ)}{q(\rmZ | \gI, \overline{\gI})} q(\rmZ | \gI, \overline{\gI}) d\rmZ \Big)          \\
                                        & =    &  & \log\Bigg( \sE_{\rmZ \sim q}\Big\{ \frac{p(\rmZ) p(\rmA | \rmZ) p(\rmS | \rmZ)}{q(\rmZ | \gI, \overline{\gI})} \Big\}\Bigg)                  \\
                                        & \geq &  & \sE_{\rmZ \sim q}\Bigg\{ \log\Big( \frac{p(\rmZ) p(\rmA | \rmZ) p(\rmS | \rmZ)}{q(\rmZ | \gI, \overline{\gI})} \Big) \Bigg\} \label{jensen},
        \end{alignat}
    }%

where (\ref{jensen}) follows from Jensen's Inequality.
Using the factorizations in \eqref{eq:q_complete_supp} and \eqref{eq:q_broken_supp}, we can now separate the factors inside $\log$ of (\ref{jensen}) as
    {\fontsize{8pt}{10pt}
        \begin{alignat}{2}
             & \sE_{\rmZ \sim q}\Bigg\{ \log\Big( \frac{p(\rmZ) p(\rmA | \rmZ) p(\rmS | \rmZ)}{q(\rmZ | \gI, \overline{\gI})} \Big) \Bigg\} \nonumber                                                        \\
             & = \sE_{\rmZ \sim q}\Bigg\{ \log\Big( \frac{p(\rmZ^{\gI}) p(\rmZ^{\overline{\gI}}) p(\rmA | \rmZ) p(\rmS | \rmZ)}{q(\rmZ^{\gI} | \gI) q(\rmZ^{\overline{\gI}} | \overline{\gI})} \Big) \Bigg\} \\
             & = \sE_{\rmZ \sim q}\Bigg\{ \log\Big( \frac{p(\rmZ^{\gI})}{ q(\rmZ^{\gI} | \gI)} \Big) + \log\Big( \frac{p(\rmZ^{\overline{\gI}})}{ q(\rmZ^{\overline{\gI}} | \overline{\gI})} \Big) \nonumber \\
             & \quad \quad \quad \quad \quad + \log\Big(p(\rmA | \rmZ)\Big) + \log\Big(p(\rmS | \rmZ)\Big) \Bigg\}                                                                                           \\
             & = - D_{KL}\Big(q(\rmZ^{\gI} | \gI) || p(\rmZ^{\gI})\Big) \nonumber                                                                                                                            \\
             & \ \ \ \ - D_{KL}\Big(q(\rmZ^{\overline{\gI}} | \overline{\gI}) || p(\rmZ^{\overline{\gI}})\Big) \nonumber                                                                                     \\
             & \ \ \ \ - \mathrm{BCE}(\Ahat, \rmA) - \mathrm{BCE}(\mathbf{\hat{S}}, \rmS)                                                                                                                    \\
             & = - D_{KL}\Big(q(\rmZ^{\gI} | \gI) || \gN(\mathbf{0}, \rmI)\Big) \nonumber                                                                                                                    \\
             & \ \ \ \ - D_{KL}\Big(q(\rmZ^{\overline{\gI}} | \overline{\gI}) || \gN(\mathbf{0}, \rmI)\Big) \nonumber                                                                                        \\
             & \ \ \ \ - \mathrm{BCE}(\Ahat, \rmA) - \mathrm{BCE}(\mathbf{\hat{S}}, \rmS)                                                                                                                    \\
             & = \gL_{\mathrm{ELBO}}  = - L_{\mathrm{recon}.}
        \end{alignat}
    }%

\section{Detailed Comparison}
We now aggregate the publicly available results for all three tasks discussed in the paper.
The publicly available approaches often cover only a subset of the datasets evaluated in this work.
So we leave the table cells empty in case of missing public results.
In addition to the competitors mentioned in the paper, we add new competitors for different tasks.

\begin{table*}[tbh]
    \caption{
        Link prediction performance, as evaluated by AUC and AP metrics.
        The best results are styled as bold and second best are underlined.
    }
    \centering
    \resizebox{1\linewidth}{!}{%
        {\setlength{\tabcolsep}{1em}%
                \begin{tabular}{@{}lrrrrrrrrrrrrrr@{}}
                    \toprule
                    \textbf{Algorithm} & \multicolumn{2}{c}{\textbf{Cora}} & \multicolumn{2}{c}{\textbf{CiteSeer}} & \multicolumn{2}{c}{\textbf{PubMed}} & \multicolumn{2}{c}{\textbf{WikiCS}} & \multicolumn{2}{c}{\textbf{CoauthorCS}} & \multicolumn{2}{c}{\textbf{AmazonComputers}} & \multicolumn{2}{c}{\textbf{AmazonPhoto}}                                                                                                                                                                                                                                                                   \\
                    \cmidrule(l{4pt}r{4pt}){2-3} \cmidrule(l{4pt}r{4pt}){4-5} \cmidrule(l{4pt}r{4pt}){6-7} \cmidrule(l{4pt}r{4pt}){8-9} \cmidrule(l{4pt}r{4pt}){10-11} \cmidrule(l{4pt}r{4pt}){12-13} \cmidrule(l{4pt}r{4pt}){14-15}
                                       & \multicolumn{1}{l}{\textbf{AUC}}  & \multicolumn{1}{l}{\textbf{AP}}       & \multicolumn{1}{l}{\textbf{AUC}}    & \multicolumn{1}{l}{\textbf{AP}}     & \multicolumn{1}{l}{\textbf{AUC}}        & \multicolumn{1}{l}{\textbf{AP}}              & \multicolumn{1}{l}{\textbf{AUC}}         & \multicolumn{1}{l}{\textbf{AP}}       & \multicolumn{1}{l}{\textbf{AUC}} & \multicolumn{1}{l}{\textbf{AP}} & \multicolumn{1}{l}{\textbf{AUC}}      & \multicolumn{1}{l}{\textbf{AP}}       & \multicolumn{1}{l}{\textbf{AUC}} & \multicolumn{1}{l}{\textbf{AP}} \\
                    \midrule
                    Spectral           & 84.60                             & 88.50                                 & 80.50                               & 85.00                               & 84.20                                   & 87.80                                        &                                          &                                       & \multicolumn{1}{l}{}             & \multicolumn{1}{l}{}            &                                       &                                       & \multicolumn{1}{l}{}             & \multicolumn{1}{l}{}            \\
                    DeepWalk           & 83.10                             & 85.00                                 & 80.50                               & 83.60                               & 84.40                                   & 84.10                                        &                                          &                                       & 91.74                            & 91.19                           & \multicolumn{1}{r}{87.35}             & \multicolumn{1}{r}{91.48}             & 91.48                            & 90.79                           \\
                    \midrule
                    GCN                & 90.47                             & 91.62                                 & 82.56                               & 83.20                               & 89.56                                   & 90.28                                        &                                          &                                       & \multicolumn{1}{l}{}             & \multicolumn{1}{l}{}            &                                       &                                       & \multicolumn{1}{l}{}             & \multicolumn{1}{l}{}            \\
                    HGCN               & 92.90                             & 93.45                                 & 95.25                               & 95.97                               & 96.30                                   & 96.75                                        &                                          &                                       & \multicolumn{1}{l}{}             & \multicolumn{1}{l}{}            &                                       &                                       & \multicolumn{1}{l}{}             & \multicolumn{1}{l}{}            \\
                    GAT                & 93.17                             & 93.81                                 & 86.48                               & 87.51                               & 91.46                                   & 92.28                                        &                                          &                                       & \multicolumn{1}{l}{}             & \multicolumn{1}{l}{}            &                                       &                                       & \multicolumn{1}{l}{}             & \multicolumn{1}{l}{}            \\
                    HGAT               & 94.02                             & 94.63                                 & 95.84                               & 95.89                               & 94.18                                   & 94.42                                        &                                          &                                       & \multicolumn{1}{l}{}             & \multicolumn{1}{l}{}            &                                       &                                       & \multicolumn{1}{l}{}             & \multicolumn{1}{l}{}            \\
                    GAE                & 91.00                             & 92.00                                 & 89.50                               & 89.90                               & 96.40                                   & 96.50                                        & \multicolumn{1}{r}{93.00}                & \multicolumn{1}{r}{94.80}             & 94.09                            & 93.86                           & \multicolumn{1}{r}{93.61}             & \multicolumn{1}{r}{93.36}             & 93.86                            & 92.96                           \\
                    VGAE               & 91.40                             & 92.60                                 & 90.80                               & 92.00                               & 94.40                                   & 94.70                                        & \multicolumn{1}{r}{93.60}                & \multicolumn{1}{r}{95.00}             & 89.60                            & 89.36                           & \multicolumn{1}{r}{93.15}             & \multicolumn{1}{r}{92.40}             & 92.05                            & 92.02                           \\
                    ARGA               & 92.40                             & 93.20                                 & 91.90                               & 93.00                               & 96.80                                   & 97.10                                        & \multicolumn{1}{r}{93.40}                & \multicolumn{1}{r}{94.70}             & 91.99                            & 92.54                           &                                       &                                       & \textbf{96.10}                   & \textbf{95.40}                  \\
                    ARVGA              & 92.40                             & 92.60                                 & 92.40                               & 93.00                               & 96.50                                   & 96.80                                        & \multicolumn{1}{r}{94.70}                & \multicolumn{1}{r}{94.80}             & 93.32                            & 93.32                           &                                       &                                       & 92.70                            & 90.90                           \\
                    DGLFRM             & 93.43                             & 93.76                                 & 93.79                               & 94.38                               & 93.95                                   & 94.97                                        &                                          &                                       & \multicolumn{1}{l}{}             & \multicolumn{1}{l}{}            &                                       &                                       & \multicolumn{1}{l}{}             & \multicolumn{1}{l}{}            \\
                    GALA               & 92.10                             & 92.20                                 & 94.40                               & 94.80                               & 91.50                                   & 89.70                                        & \multicolumn{1}{r}{93.60}                & \multicolumn{1}{r}{93.10}             & 93.81                            & 94.49                           &                                       &                                       & 91.80                            & 91.00                           \\
                    Graphite           & 94.70                             & 94.90                                 & 97.30                               & 97.40                               & 97.40                                   & 97.40                                        &                                          &                                       & \multicolumn{1}{l}{}             & \multicolumn{1}{l}{}            &                                       &                                       & \multicolumn{1}{l}{}             & \multicolumn{1}{l}{}            \\
                    \midrule
                    DGI                & 89.80                             & 89.70                                 & 95.50                               & 95.70                               & 91.20                                   & 92.20                                        &                                          &                                       & 94.87                            & 94.34                           &                                       &                                       & 92.24                            & 92.14                           \\
                    GIC                & 93.50                             & 93.30                                 & 97.00                               & 96.80                               & 93.70                                   & 93.50                                        &                                          &                                       & 95.03                            & 94.94                           &                                       &                                       & 92.70                            & 92.34                           \\
                    GMI                & 95.10                             & 95.60                                 & 97.80                               & 97.40                               & 96.37                                   & 96.04                                        &                                          &                                       & \underline{96.37}                & 95.04                           &                                       &                                       & 93.88                            & 92.67                           \\
                    GCA                & 95.75                             & 95.47                                 & 96.44                               & 96.49                               & 95.28                                   & 95.52                                        &                                          &                                       & 96.31                            & \textbf{96.28}                  &                                       &                                       & 93.25                            & 92.74                           \\
                    MVGRL              & 90.52                             & 90.45                                 & 92.89                               & 92.89                               & 92.45                                   & 92.17                                        &                                          &                                       & 95.17                            & 95.58                           &                                       &                                       & 92.89                            & 92.45                           \\
                    \midrule
                    G-BT               & 87.46                             & 86.84                                 & 93.42                               & 93.01                               & 94.53                                   & 94.26                                        & \multicolumn{1}{r}{93.05}                & \multicolumn{1}{r}{93.18}             & 92.64                            & 91.40                           & \multicolumn{1}{r}{91.54}             & \multicolumn{1}{r}{90.59}             & 95.12                            & 94.45                           \\
                    \midrule
                    BGAE               & \underline{98.52}                 & \underline{98.42}                     & \underline{98.59}                   & \textbf{98.61}                      & 97.78                                   & 97.68                                        & \multicolumn{1}{r}{97.23}                & \multicolumn{1}{r}{97.69}             & 96.31                            & 95.44                           & \multicolumn{1}{r}{\underline{95.01}} & \multicolumn{1}{r}{94.04}             & 95.01                            & 94.24                           \\
                    BGAE + Att         & \textbf{98.79}                    & \textbf{98.73}                        & 98.56                               & \underline{98.57}                   & \textbf{98.06}                          & \textbf{98.03}                               & \multicolumn{1}{r}{\underline{97.73}}    & \multicolumn{1}{r}{\underline{98.03}} & \textbf{96.51}                   & \underline{95.65}               & \multicolumn{1}{r}{\textbf{95.13}}    & \multicolumn{1}{r}{\textbf{94.48}}    & \underline{95.18}                & \underline{94.49}               \\
                    BVGAE              & 97.87                             & 97.62                                 & \textbf{98.63}                      & \underline{98.57}                   & \underline{97.93}                       & \underline{97.89}                            & \multicolumn{1}{r}{97.62}                & \multicolumn{1}{r}{97.42}             & 96.12                            & 95.13                           & \multicolumn{1}{r}{94.66}             & \multicolumn{1}{r}{94.01}             & 94.61                            & 94.29                           \\
                    BVGAE + Att        & 98.03                             & 97.77                                 & 98.23                               & 98.08                               & 97.77                                   & 97.74                                        & \multicolumn{1}{r}{\textbf{97.85}}       & \multicolumn{1}{r}{\textbf{98.05}}    & 96.21                            & 95.34                           & \multicolumn{1}{r}{94.90}             & \multicolumn{1}{r}{\underline{94.07}} & 94.97                            & 94.28                           \\
                    \bottomrule
                \end{tabular}
            }
    }\label{tab:link-prediction-complete}
\end{table*}

\subsection{Link Prediction} \label{supp:link_prediction}
We add the following approaches in addition to the ones given in \tabref{tab:link-prediction}:
GNN based architectures \textbf{GCN}~\cite{gcn} and \textbf{GAT}~\cite{gat}, along with their hyperbolic variants \textbf{HGCN}~\cite{hgcn} and \textbf{HGAT}~\cite{hgat}.
Deep Generative Latent Feature Relational Model or \textbf{DGLFRM}~\cite{dglfrm} that aims to reconstruct the adjacency matrix while retaining the interpretability of stochastic block models.
Graph InfoClust or \textbf{GIC}~\cite{gic}, which learns network embedding by maximizing the mutual information with respect to the graph-level summary as well as the cluster-level summaries.
\textbf{Graphite}~\cite{graphite}, which is another auto-encoder based generative model that employs a multi-layer procedure, inspired by low-rank approximations, to iteratively refine the reconstructed graph via message passing.

\subsubsection{Results}
We can notice that our approach is still either best or second best including all the competitors.
GMI and GCA achieve good results for CoauthorCS, Graphite performs consistently well for all the datasets.
However these algorithms suffer when evaluated for clustering and transductive node classification.

\subsection{Clustering} \label{supp:clustering}

\begin{table*}[tbh]
    \captionsetup{width=1\linewidth}
    \caption{
        Node clustering performance, as evaluated by NMI.
        The best results are styled as bold and second best are underlined.
        OOM refers to Out-of-Memory.
    }
    \centering
    \resizebox{1\linewidth}{!}{%
        {\setlength{\tabcolsep}{0.7em}%
                \begin{tabular}{lrrrrrrrr}
                    \toprule
                    \textbf{Algorithm} & \textbf{Cora}        & \textbf{CiteSeer}    & \textbf{PubMed}      & \textbf{WikiCS}                       & \textbf{CoauthorCS}                   & \textbf{CoauthorPhysics}              & \textbf{AmazonComputers}              & \textbf{AmazonPhoto} \\
                    \midrule
                    K-means            & 32.10                & 30.50                & 0.10                 & \multicolumn{1}{r}{18.20}             & \multicolumn{1}{r}{64.20}             & \multicolumn{1}{r}{48.90}             & \multicolumn{1}{r}{16.60}             & 28.20                \\
                    Spectral           & 12.70                & 5.60                 & 4.20                 &                                       &                                       &                                       &                                       & \multicolumn{1}{l}{} \\
                    \midrule
                    DeepWalk           & 32.70                & 8.80                 & 27.90                &                                       &                                       &                                       &                                       & \multicolumn{1}{l}{} \\
                    BigClam            & 0.70                 & 3.60                 & 0.60                 &                                       &                                       &                                       &                                       & \multicolumn{1}{l}{} \\
                    DNGR               & 31.80                & 18.00                & 15.50                &                                       &                                       &                                       &                                       & \multicolumn{1}{l}{} \\
                    RMSC               & 25.50                & 13.90                & 25.50                &                                       &                                       &                                       &                                       & \multicolumn{1}{l}{} \\
                    TADW               & 44.10                & 29.10                & 0.10                 &                                       &                                       &                                       &                                       & \multicolumn{1}{l}{} \\
                    \midrule
                    GAE                & 42.90                & 17.60                & 27.70                & \multicolumn{1}{r}{24.30}             & \multicolumn{1}{r}{73.10}             & \multicolumn{1}{r}{54.50}             & \multicolumn{1}{r}{44.10}             & 61.60                \\
                    VGAE               & 43.60                & 15.60                & 22.90                & \multicolumn{1}{r}{26.10}             & \multicolumn{1}{r}{73.30}             & \multicolumn{1}{r}{56.30}             & \multicolumn{1}{r}{42.30}             & 53.00                \\
                    ARGA               & 44.90                & 35.00                & 30.50                & \multicolumn{1}{r}{27.50}             & \multicolumn{1}{r}{66.80}             & \multicolumn{1}{r}{51.20}             & \multicolumn{1}{r}{23.50}             & 57.70                \\
                    ARVGA              & 52.60                & 33.80                & 29.00                & \multicolumn{1}{r}{28.70}             & \multicolumn{1}{r}{61.60}             & \multicolumn{1}{r}{52.60}             & \multicolumn{1}{r}{23.70}             & 45.50                \\
                    GALA               & 57.70                & \underline{44.10}    & 32.70                &                                       &                                       &                                       &                                       & 51.20                \\
                    Graphite           & 54.12                & 42.42                & 32.40                &                                       &                                       &                                       &                                       & \multicolumn{1}{l}{} \\
                    \midrule
                    DGI                & 41.10                & 31.50                & 27.70                & \multicolumn{1}{r}{31.00}             & \multicolumn{1}{r}{74.70}             & \multicolumn{1}{r}{67.00}             & \multicolumn{1}{r}{31.80}             & 37.60                \\
                    GIC                & 53.70                & \textbf{45.30}       & 31.90                &                                       &                                       &                                       &                                       & \multicolumn{1}{l}{} \\
                    GRACE              & 46.18                & 38.29                & 16.27                & \multicolumn{1}{r}{42.82}             & \multicolumn{1}{r}{75.62}             & OOM                                   & \multicolumn{1}{r}{47.93}             & 65.13                \\
                    AGC                & 53.70                & 41.10                & 31.60                &                                       &                                       &                                       &                                       & \multicolumn{1}{l}{} \\
                    GMI                & 50.33                & 38.14                & 26.20                &                                       &                                       &                                       &                                       & \multicolumn{1}{l}{} \\
                    GMNN               & 53.72                & 41.73                & 31.77                &                                       &                                       &                                       &                                       & \multicolumn{1}{l}{} \\
                    DAEGC              & 52.80                & 39.70                & 26.60                &                                       &                                       &                                       &                                       & \multicolumn{1}{l}{} \\
                    DBGAN              & 56.00                & 40.70                & 32.40                &                                       &                                       &                                       &                                       & 48.50                \\
                    GCA                & 55.70                & 37.40                & 28.90                & \multicolumn{1}{r}{29.90}             & \multicolumn{1}{r}{73.50}             & \multicolumn{1}{r}{59.40}             & \multicolumn{1}{r}{42.60}             & 34.40                \\
                    MVGRL              & 60.90                & 44.00                & 31.50                & \multicolumn{1}{r}{26.30}             & \multicolumn{1}{r}{74.00}             & \multicolumn{1}{r}{59.40}             & \multicolumn{1}{r}{24.40}             & 34.40                \\
                    BGRL               & \multicolumn{1}{l}{} & \multicolumn{1}{l}{} & \multicolumn{1}{l}{} & \multicolumn{1}{r}{39.69}             & \multicolumn{1}{r}{77.32}             & \multicolumn{1}{r}{55.68}             & \multicolumn{1}{r}{53.64}             & \textbf{68.41}       \\
                    \midrule
                    G-BT               & 43.40                & 41.57                & 29.52                & \multicolumn{1}{r}{27.46}             & \multicolumn{1}{r}{74.37}             & \multicolumn{1}{r}{59.80}             & \multicolumn{1}{r}{65.55}             & 52.39                \\
                    \midrule
                    BGAE               & \textbf{62.42}       & 43.36                & \underline{38.46}    & \multicolumn{1}{r}{\underline{45.80}} & \multicolumn{1}{r}{\underline{80.10}} & \multicolumn{1}{r}{\underline{68.01}} & \multicolumn{1}{r}{\textbf{66.98}}    & 67.13                \\
                    BGAE + Att         & \underline{62.27}    & 43.84                & \textbf{38.59}       & \multicolumn{1}{r}{\textbf{46.93}}    & \multicolumn{1}{r}{\textbf{80.30}}    & \multicolumn{1}{r}{\textbf{68.12}}    & \multicolumn{1}{r}{\underline{66.93}} & \underline{67.43}    \\
                    BVGAE              & 59.60                & 43.29                & 37.41                & \multicolumn{1}{r}{40.78}             & \multicolumn{1}{r}{79.01}             & \multicolumn{1}{r}{67.10}             & \multicolumn{1}{r}{60.98}             & 61.33                \\
                    BVGAE + Att        & 59.82                & 43.27                & 37.47                & \multicolumn{1}{r}{40.86}             & \multicolumn{1}{r}{79.42}             & \multicolumn{1}{r}{67.06}             & \multicolumn{1}{r}{61.44}             & 61.62                \\
                    \bottomrule
                \end{tabular}
            }
    }\label{tab:clustering-complete}
\end{table*}
We include the following clustering-specific competitors in addition to the ones given in \tabref{tab:clustering} and \secref{supp:link_prediction}:
\textbf{BigClam}~\cite{bigclam} uses matrix factorization for community detection.
\textbf{DNGR}~\cite{dngr} learns network embedding by using stacked denoising auto-encoders.
\textbf{RMSC}~\cite{rmsc} introduces a multi-view spectral clustering approach to recover a low-rank transition probability matrix from the transition matrices corresponding to multiple views of input data.
\textbf{TADW}~\cite{tadw} learns the network embedding by treating DeepWalk as matrix factorization and adding the features of vertices.
\textbf{AGC}~\cite{agc} performs attributed graph clustering by first obtaining smooth node feature representations via k-order graph convolution and then performing spectral clustering on the learned features.
\textbf{DAEGC}~\cite{daegc} uses GAT to encode the importance of the neighboring nodes in the latent space such that both the reconstruction loss and the KL-divergence based clustering loss are minimized.

In addition, we include some well-known approaches for unsupervised network embedding.
\textbf{DBGAN}~\cite{dbgan} introduces a bidirectional adversarial learning framework to learn network embedding in such a way that the prior distribution is also estimated along with the adversarial learning.
\textbf{GMI}~\cite{gmi} is an unsupervised approach to learn node representations while aiming to improve generalization performance via added contrastive regularization.
\textbf{GMNN}~\cite{gmnn} relies on a random field model, which can be trained with variational expectation maximization.

\subsubsection{Results}
Our approach performs the best overall, although GIC and BGRL achieve the best results for CiteSeer and AmazonPhoto respectively.
Graphite performed well in \tabref{tab:link-prediction-complete}, but for clustering, it is outperformed by many other competitors.
The converse is true for GIC and GALA, which outperform \ourss for a single dataset in clustering, but fail to compete in the link-prediction task.
Similarly, DBGAN emerges as a decent competitor for node-clustering.
However, as we will see in the next section, its performance degrades for the task of node classification.
We cannot comment on BGRL because we could neither find its public implementation nor any publicly published results for link-prediction on the selected datasets.

\subsection{Transductive Node Classification}
\begin{table*}[tbh]
    \captionsetup{width=1\linewidth}
    \caption{
        Transductive node classification performance, as evaluated by accuracy.
        The best results are styled as bold and second best are underlined.
    }
    \centering
    \resizebox{1\linewidth}{!}{%
        {\setlength{\tabcolsep}{1em}%
                \begin{tabular}{@{}lrrrrrrrr@{}}
                    \toprule
                    \textbf{Algorithm} & Cora              & CiteSeer          & PubMed            & WikiCS            & CoauthorCS     & CoauthorPhysics   & AmazonComputers   & AmazonPhoto       \\
                    \midrule
                    Raw                & 47.87             & 49.33             & 69.11             & 71.98             & 90.37          & 93.58             & 73.81             & 78.53             \\
                    DeepWalk           & 70.66             & 51.39             & 74.31             & 77.21             & 87.70          & 94.90             & 86.28             & 90.05             \\
                    \midrule
                    GAE                & 71.53             & 65.77             & 72.14             & 70.15             & 90.01          & 94.92             & 85.18             & 91.68             \\
                    VGAE               & 75.24             & 69.05             & 75.29             & 75.63             & 92.11          & 94.52             & 86.44             & 92.24             \\
                    ARGA               & 74.14             & 64.14             & 74.12             & 66.88             & 89.41          & 93.10             & 84.39             & \underline{92.68} \\
                    ARVGA              & 74.38             & 64.24             & 74.69             & 67.37             & 88.54          & 94.30             & 84.66             & 92.49             \\
                    Graphite           & 82.10             & 71.00             & 79.30             &                   &                &                   &                   &                   \\
                    \midrule
                    DGI                & 81.68             & 71.47             & 77.27             & 75.35             & 92.15          & 94.51             & 83.95             & 91.61             \\
                    GIC                & 81.73             & 71.93             & 77.33             & 77.28             & 89.40          & 93.10             & 84.89             & 92.11             \\
                    GRACE              & 80.04             & 71.68             & 79.53             & \textbf{80.14}    & 92.51          & 94.70             & 87.46             & 92.15             \\
                    AGC                & 70.90             & 71.89             & 68.91             &                   &                &                   &                   &                   \\
                    GMI                & 83.05             & \underline{73.03} & 80.10             & 74.85             & OOM            & OOM               & 82.21             & 90.68             \\
                    GMNN               & 82.78             & 71.54             & 80.60             &                   &                &                   &                   &                   \\
                    DBGAN              & 77.30             & 69.70             & 77.50             &                   &                &                   &                   &                   \\
                    GCA                & 82.10             & 71.30             & 80.20             & 78.23             & 92.95          & \textbf{95.73}    & 88.94             & 92.53             \\
                    MVGRL              & 82.90             & 72.60             & 79.40             & 77.52             & 92.11          & 92.11             & 87.52             & 91.74             \\
                    BGRL               & 82.70             & 71.10             & 79.60             & \underline{79.98} & 93.31          & 95.56             & 89.68             & \textbf{92.87}    \\
                    \midrule
                    G-BT               & 80.80             & 73.00             & 80.00             & 76.65             & 92.95          & 95.07             & 88.14             & 92.63             \\
                    \midrule
                    BGAE               & \underline{83.51} & 72.43             & \textbf{81.84}    & 78.93             & \textbf{93.76} & 95.01             & \underline{92.24} & 91.10             \\
                    BGAE + Att         & \textbf{83.60}    & 72.41             & \underline{80.95} & 79.53             & \textbf{93.76} & \underline{95.64} & \textbf{92.44}    & 91.89             \\
                    BVGAE              & 82.62             & 72.97             & 80.02             & 77.52             & 93.25          & 95.13             & 89.19             & 89.38             \\
                    BVGAE + Att        & 82.57             & \textbf{73.09}    & 80.25             & 77.82             & 93.15          & 95.60             & 89.91             & 89.98             \\
                    \bottomrule
                \end{tabular}
            }
    }\label{tab:classification-complete}
\end{table*}

The competitors evaluated in \tabref{tab:classification-complete} have already been introduced in the main paper and in \secref{supp:link_prediction} and \secref{supp:clustering}.
We exclude some methods that are specifically designed for clustering, because such methods perform poor on transductive node classification.
\subsubsection{Results}
GRACE, GMI, GALA, and BGRL perform well for WikiCS, CiteSeer, and AmazonPhoto.
However our approach performs the best overall as we achieve the best or second-best results in 6 out of 8 datasets.
This demonstrates the efficacy of our approach over a variety of tasks unlike many competitors that shine only in some of the target tasks.

\section{Ablation Studies}
We now observe how our approach is affected by changes in $\beta$ from \eqref{eq:general-loss}, $\lambda$ from \eqref{eq:l_cov}, and average-node degree used for sparsification of $\overline{\rmS}$ into $\rmS$.
For sake of brevity, we only plot the results for transductive node classification task because link prediction and clustering follow a similar pattern.
To emphasize the relative performance, the vertical axes correspond to the percentage accuracy scores relative to the ones reported in \tabref{tab:classification}.

\subsection{Effect of $\beta$}
To evaluate the effect of $\beta$ in \eqref{eq:general-loss}, we sweep $\beta$ for the values across the set $\{0, 0.1, 1, 10, 100, 1000, 10000\}$.
\begin{figure}[tbh]
    \centering
    \caption{
        Effect of $\beta$ on transductive node classification performance.
        The vertical axis shows the performance in \%, relative to the results reported in the main paper.
    }
    \includegraphics[width=1\linewidth]{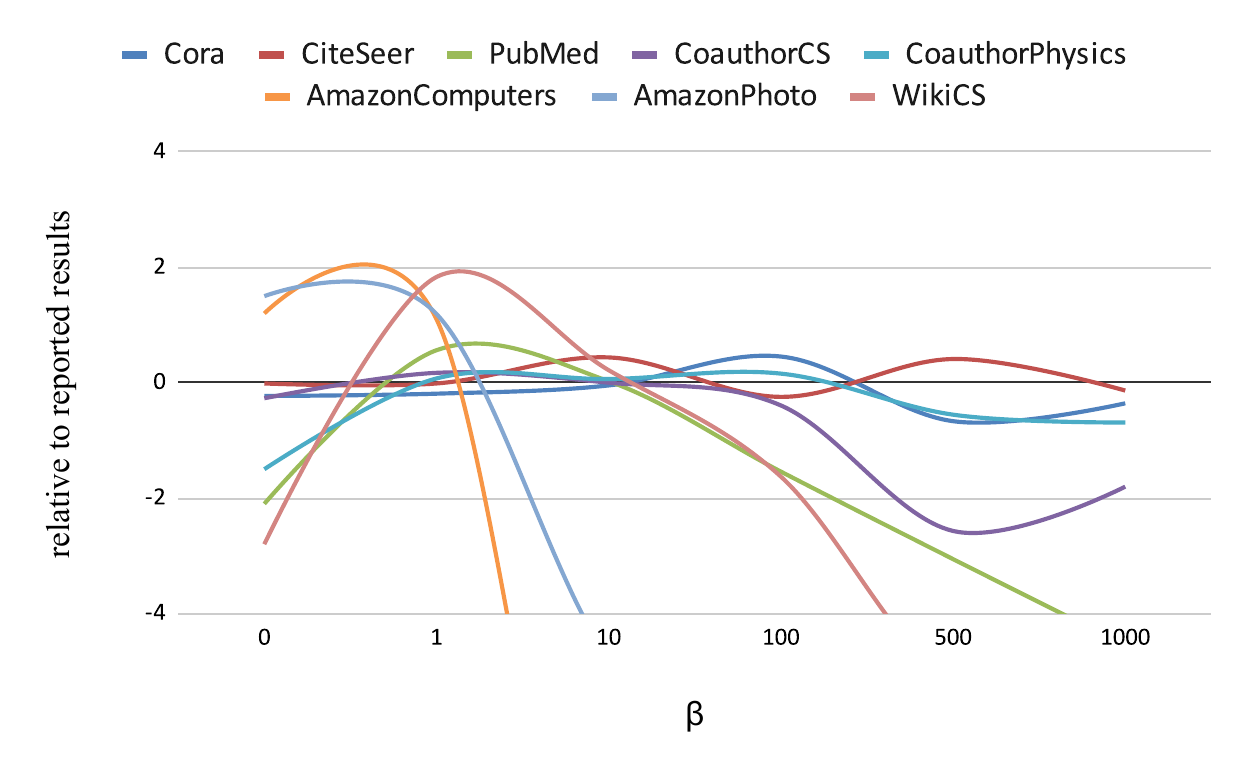}
\end{figure} \label{fig:ablation:beta}

\Figref{fig:ablation:beta} shows the effect of $\beta$ on transductive node classification for different datasets.
For most of the datasets, the results are rather stable for quite a large range of $\beta$, i.e., in [1, 50] range.
AmazonComputers and AmazonPhoto datasets are an exception in the sense that their performance degrades quicker than other datasets.
Overall, a general trend of degradation can be observed for high values of $\beta$ for all datasets, which is intuitive because for such values, the covariance loss takes over and the reconstruction loss is practically neglected, resulting in relatively poor results.
Another observation is that the results above the 0-line on the graphs are better than the ones reported in the main paper.
So, by carefully tuning $\beta$, we can achieve even better results compared to the ones reported in the paper.

\subsection{Effect of $\lambda$}
The hyperparameter $\lambda$ governs the trade-off between invariance and cross-covariance in \eqref{eq:l_cov}.
The proposed value of $\lambda$ in \cite{barlow} is $5e^{-3}$.
To see the effect of changing $\beta$, we sweep it across the values $\{1e^{-3}, 5e^{-3}, 1e^{-2}, 5e^{-2}\}$.
The effect of changing $\lambda$ on different datasets has been plotted in \figref{fig:ablation:lambda}.
The plot validates that $\lambda = 5e^{-3}$, proposed in Barlow Twins\cite{barlow}, is a reasonable choice also for graph datasets.
Some datasets perform better for $\lambda = 1e^{-3}$ and some yield better results for $\lambda = 1e^{-2}$.
However, there is a general trend of decrease in the performance for $\lambda \geq 5e^{-2}$.

\begin{figure}[tbh]
    \centering
    \caption{
        Effect of $\lambda$ on transductive node classification performance.
        The vertical axis shows the performance in \%, relative to the results reported in the main paper.
    }
    \includegraphics[width=1\linewidth]{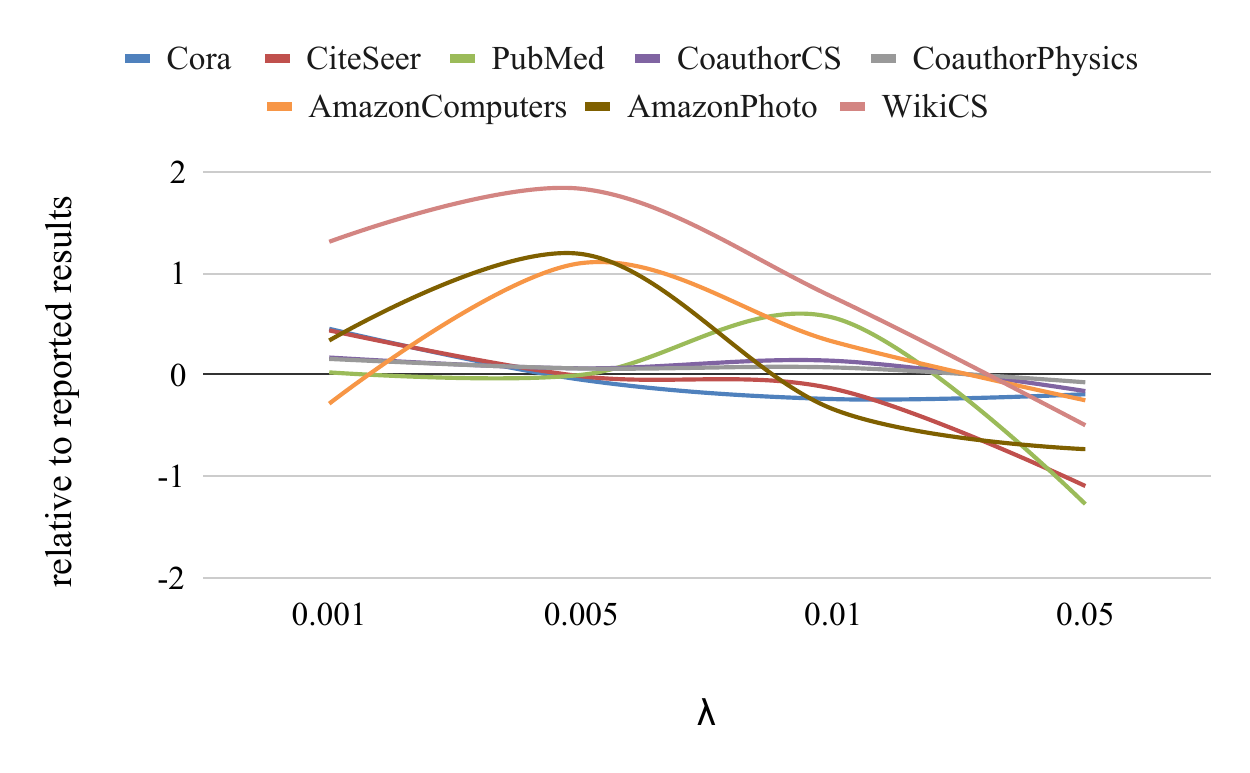}
\end{figure} \label{fig:ablation:lambda}

\subsection{Effect of Average Sparsification Degree in $\rmS$}
As mentioned in \secref{sec:gdc}, one of the ways of sparsification is to provide the intended average node degree.
We have fixed this value to 25 for all the reported results.
Now we observe the effect of changing this hyperparameter.
For this purpose, we sweep the average degree over the values $\{5, 10, 20, 25, 30, 40, 50, 60, 75, 80, 100, 125, 150\}$.
The results have been plotted in \figref{fig:ablation:average_degree}.
The results for PubMed are not plotted for the degree values greater than 75 because of out-of-memory issues.
The relative performance remains more or less consistent over the plotted range, and varies between $\pm 1\%$ of the reported results.
This also shows that the architecture can extract the relevant information from the neighborhood over a reasonable range of average degree.
An exception is WikiCS where the results improve by up to 2\% compared to the reported results in \tabref{tab:classification} for the average degree value of 150.
However, for such a high value, the graph is no longer reasonably sparse.
This causes high training overhead because of the large number of edges in $\rmS$.
On the other extreme, for the value of 5, we can see a decline in many datasets because here $\rmS$ is too sparse, hence the information in $\rmS$ is too little to be of use.
\begin{figure}[tbh]
    \centering
    \caption{
        Effect of average node degree in $\rmS$ on transductive node classification performance.
        The vertical axis shows the performance in \%, relative to the results reported in the main paper.
    }
    \includegraphics[width=1\linewidth]{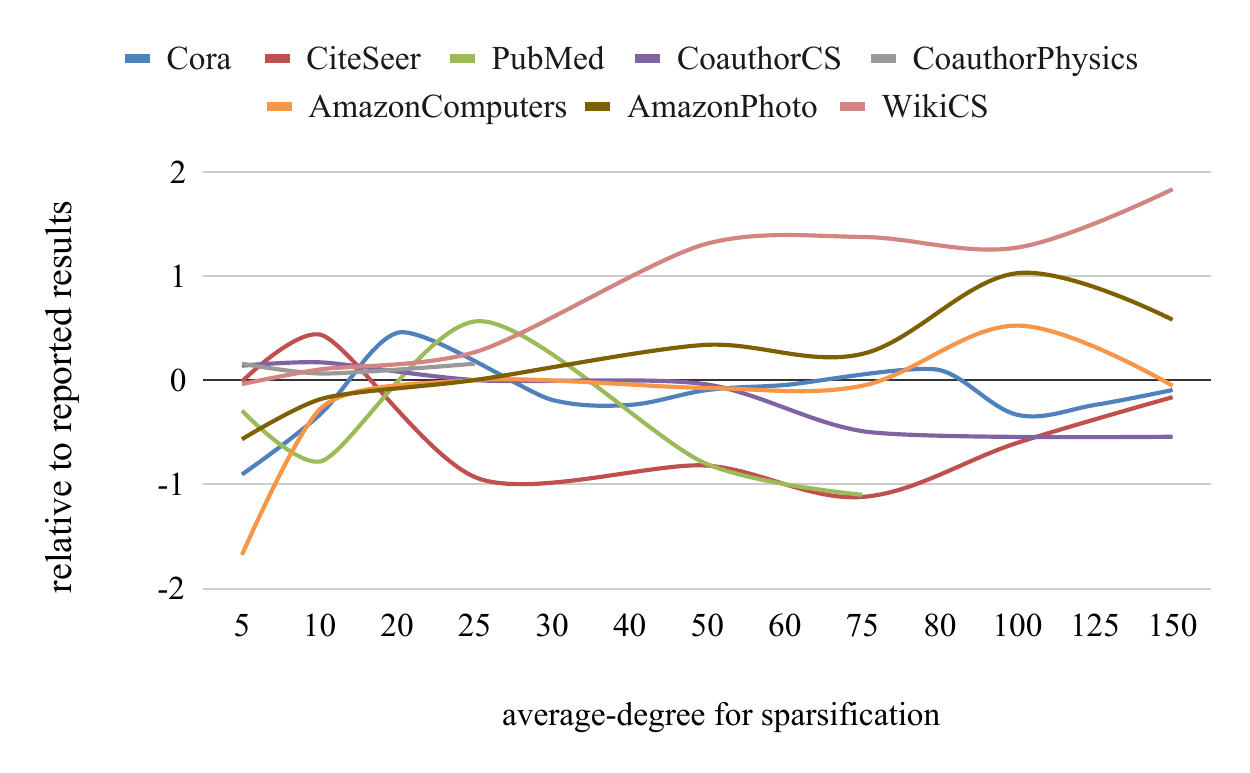}
\end{figure} \label{fig:ablation:average_degree}

\section{Variants of Our Approach}
In the main paper, we have reported the results for PPR both with and without attention.
From these results, we can already establish that it is always better to use attention i.e., let the neural network decide the weights for averaging the embeddings from the immediate and larger neighborhood.
So, in this section, we focus on the case with attention, and report the results with following variations:
\begin{itemize}
    \item Toggling $L_{\mathrm{recon}}$ on/off in \eqref{eq:general-loss}
    \item Choosing between \ourss and \vours.
    \item Choosing between PPR and Heat Kernels for diffusion.
\end{itemize}

\begin{figure*}[tbh]
    \centering
    \begin{subfigure}[t]{1\linewidth}
        \centering
        \includegraphics[width=1\linewidth]{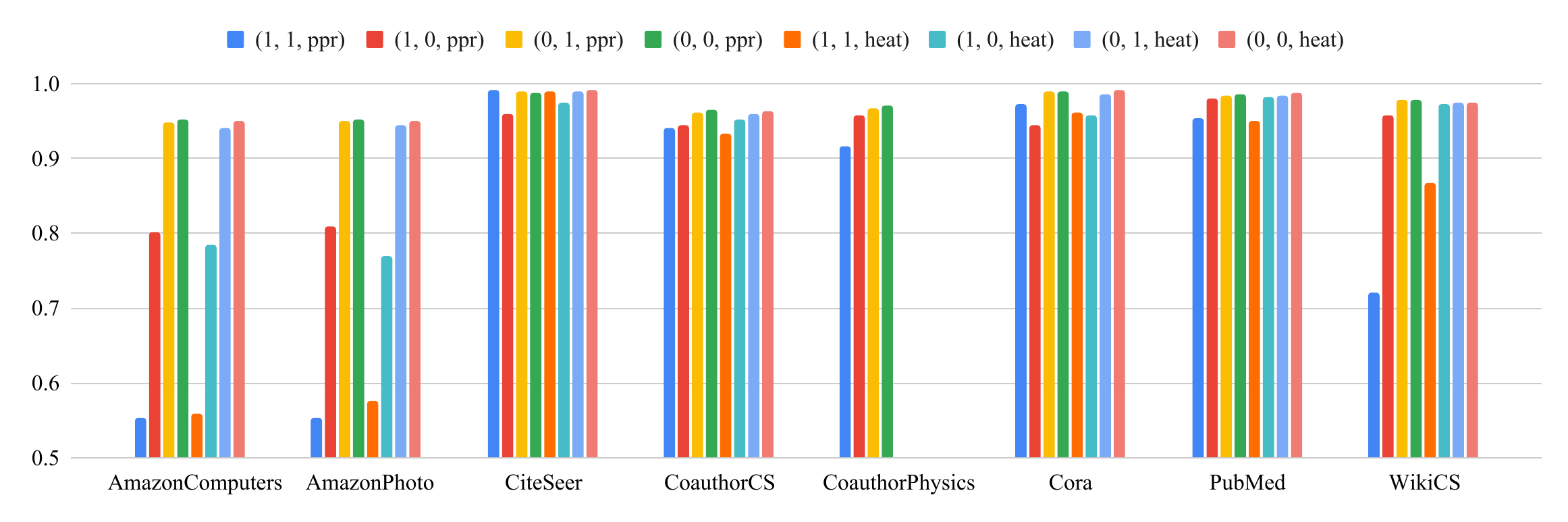}
        \caption{Variants of our approach for link prediction}
        \label{fig:variants:link-prediction}
    \end{subfigure}
    \begin{subfigure}[t]{1\linewidth}
        \centering
        \includegraphics[width=1\linewidth]{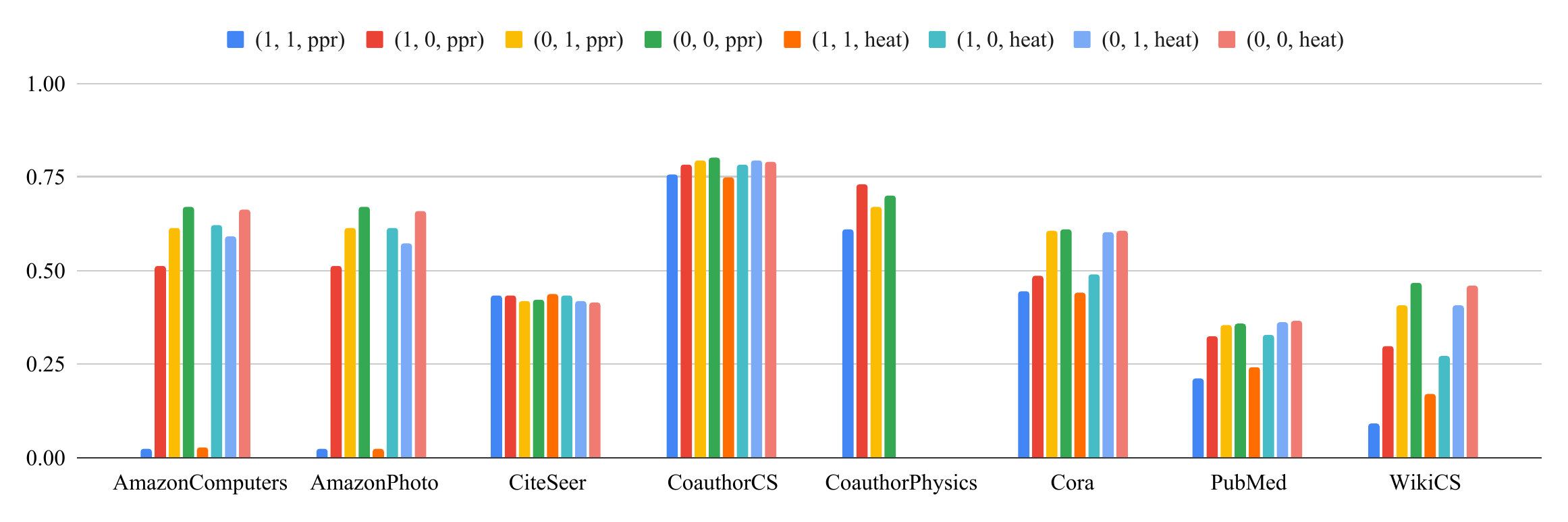}
        \caption{Variants of our approach for node clustering}
        \label{fig:variants:clustering}
    \end{subfigure}
    \begin{subfigure}[t]{1\linewidth}
        \centering
        \includegraphics[width=1\linewidth]{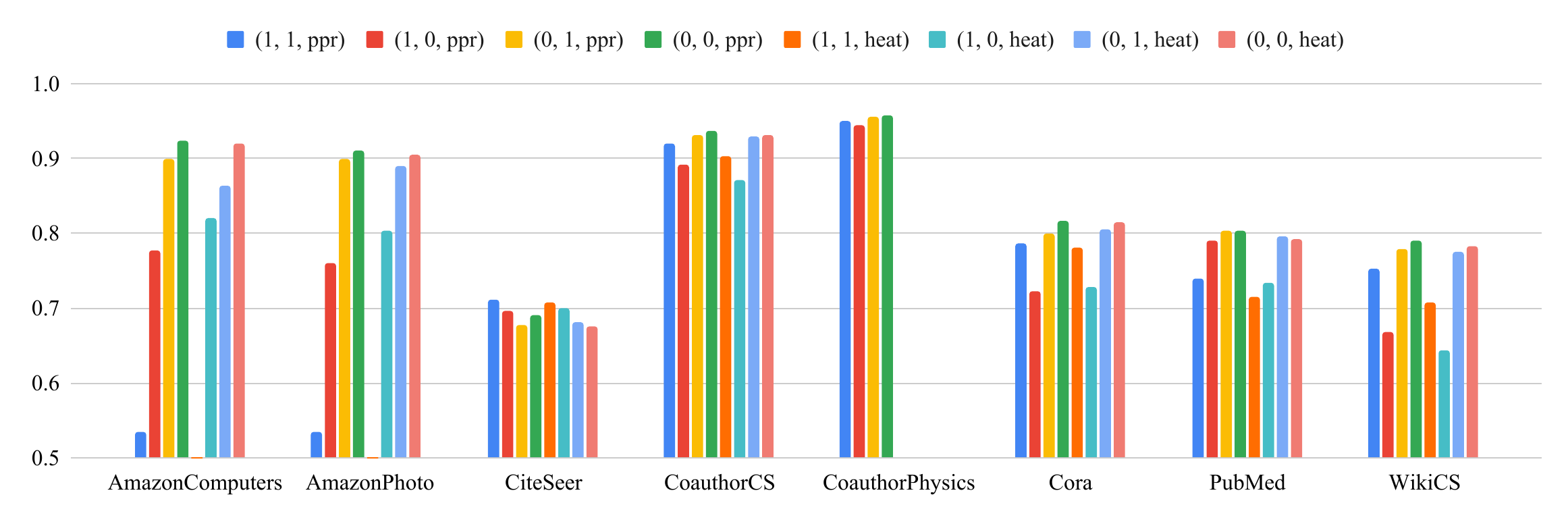}
        \caption{Variants of our approach for node classification}
        \label{fig:variants:classification}
    \end{subfigure}
    \caption{Variants of our approach for link prediction, node clustering, and node classification.}
    \label{fig:variants:main}
\end{figure*}

For brevity, we use triple of the form $\Big(\mathbf{1}(L_{\mathrm{recon}} \text{ is mute}), \mathbf{1}( \text{ variational model} ), \text{kernel name}\Big)$.
For instance, (1, 0, ppr) means that we are referring to the variant where we are only using $L_{\mathrm{cov}}$ in non-variational mode with PPR kernel for diffusion.
In the main paper, we have reported the results for the variant (0, 0, ppr) and (0, 1, ppr).
Using this notation, we plot the results for all the eight variations for all three tasks i.e. link prediction, clustering and transductive node classification in \figref{fig:variants:link-prediction}, \figref{fig:variants:clustering}, and \figref{fig:variants:classification} respectively.
For some datasets (e.g., CoauthorPhysics), some variants could not be plotted because of out-of-memory issues.
The general behavior is similar for different variants across all three tasks.
The important observations from \figref{fig:variants:main} are as follows:
\begin{itemize}
    \item The variant (0, 0, ppr), shown in green, performs the best overall.
    \item The variants (0, 0, ppr) and (0, 0, heat) are usually close in performance, although (0, 0, ppr) is often better by a small margin.
    \item The variants (0, 0, ppr) and (0, 0, heat) with simple GCN encoders usually outperform their variational counterparts, i.e., (0, 1, ppr) and (0, 1, heat).
          There are, however, minor exceptions.
          For instance, in \figref{fig:variants:link-prediction}, (0, 1, ppr) is marginally better than (0, 0, ppr) for CiteSeer.
          Similarly, in \figref{fig:variants:clustering}, (0, 1, heat) is marginally better than (0, 0, heat) for CoauthorCS.
    \item When $L_{\mathrm{recon}}$ is turned off, the performance is usually relatively worse than when $L_{\mathrm{recon}}$ is on.
          This can be seen in (1, 1, ppr), (1, 0, ppr), (1, 1, heat), and (1, 0, heat) variants.
          The only exception is CiteSeer in \figref{fig:variants:link-prediction} where (1, 1, ppr) outperforms (0, 1, ppr) by a tiny margin.
          This validates our intuition that $L_{\mathrm{recon}}$ aids $L_{\mathrm{cov}}$ almost always.
\end{itemize}

\end{document}